\documentclass{article} 
\usepackage{iclr2019_conference,times}

\usepackage{amsmath,amsfonts,bm}

\def\eqref#1{equation~\ref{#1}}

\def\1{\bm{1}}

\DeclareMathAlphabet{\mathsfit}{\encodingdefault}{\sfdefault}{m}{sl}
\SetMathAlphabet{\mathsfit}{bold}{\encodingdefault}{\sfdefault}{bx}{n}

\usepackage{hyperref}
\usepackage{url}

\usepackage{times}  
\usepackage{helvet}  
\usepackage{courier}  
\usepackage{url}  
\usepackage{graphicx}  

\usepackage{algorithm}
\usepackage[noend]{algpseudocode}

\usepackage{eufrak}
\usepackage{subcaption}
\usepackage[font=small,labelfont=bf]{caption}
\usepackage[font=small,labelfont=bf]{subcaption}
\usepackage{amsmath}
\usepackage{pgfplots}
\usepackage{tikz}
\usepackage{multirow}

\usepackage{arydshln}

\include{plots}
\definecolor{bblue}{HTML}{4F81BD}
\definecolor{rred}{HTML}{C0504D}
\definecolor{ggreen}{HTML}{9BBB59}
\definecolor{ppurple}{HTML}{9F4C7C}
\definecolor{Dark scarlet}{HTML}{560319}
\definecolor{Forest green}{HTML}{1E4D2B}

\title{
{\it textTOvec}: Deep Contextualized Neural \\Autoregressive Topic Models of Language\\ with Distributed Compositional Prior 
}

\iclrfinalcopy  

\newcommand*{\affaddr}[1]{#1} 
\newcommand*{\affmark}[1][*]{\textsuperscript{#1}}

\author{Pankaj Gupta\affmark[1,2], Yatin Chaudhary\affmark[1], Florian Buettner\affmark[1], Hinrich Sch\"{u}tze\affmark[2]\\ 
 \affaddr{\affmark[1]Corporate Technology, Machine-Intelligence (MIC-DE), Siemens AG  Munich, Germany}\\
  \affaddr{\affmark[2]CIS, University of Munich (LMU) Munich, Germany} \\
  {\tt \{pankaj.gupta, yatin.chaudhary, buettner.florian\}@siemens.com}\\
}

\newcounter{notecounter}
\newcommand{\enotesoff}{\long\gdef\enote##1##2{}}

\enotesoff

\begin{document}

\maketitle

\begin{abstract}

We address two challenges of probabilistic topic modelling in order to better estimate the probability of a word in a given context, i.e., $P (\mbox{word} | \mbox{context})$ 
: 
(1) {\it No language structure in context}:  Probabilistic topic models ignore word order by summarizing a given context as a ``bag-of-word"  and consequently  
the semantics of words in the context is lost. 
In this work, we incorporate language structure by combining a neural autoregressive topic model (TM) (e.g., DocNADE) with a LSTM based language model (LSTM-LM) in a single probabilistic framework. 
The LSTM-LM learns a vector-space representation of each word by accounting for word order in local collocation patterns, while the TM simultaneously learns a latent representation from the entire document.  
In addition, the LSTM-LM models complex characteristics of language (e.g., syntax and semantics), while the TM discovers the underlying thematic structure in a collection of documents.  
We unite two complementary paradigms of learning the meaning of word occurrences by combining a topic model and a language model in a unified probabilistic framework, named as ctx-DocNADE. 
(2) {\it Limited context and/or smaller training corpus of documents}:  In settings with a small
number of word occurrences (i.e., lack of context) in short
text or data sparsity in a corpus of few documents, the application
of TMs is challenging. 
We address this challenge by incorporating external knowledge into neural autoregressive topic models via a language modelling approach: we use word embeddings as input of a LSTM-LM 
with the aim to improve the word-topic mapping on a smaller and/or short-text corpus. The proposed DocNADE extension is named as ctx-DocNADEe.

We present novel neural autoregressive topic model variants coupled with neural language models and embeddings priors that consistently outperform state-of-the-art generative topic 
models in terms of generalization (perplexity), interpretability (topic coherence)
and applicability (retrieval and classification) over 7 
long-text and 8 short-text datasets from diverse domains.

\end{abstract}

\section{Introduction}\label{introduction}

Probabilistic topic models, such as LDA \citep{Blei:81}, Replicated Softmax (RSM) \citep{Salakhutdinov:82} and Document  Neural Autoregressive Distribution Estimator (DocNADE) variants \citep{Hugo:82,Hugo:83,HugoJMLR:82, pankajgupta:2019iDocNADEe}  
are often used to extract topics from text collections, and predict the probabilities of each word in a given document belonging to each topic.  
Subsequently, they learn latent document representations that can be used to perform natural language processing  (NLP) tasks
such as information retrieval (IR), document classification or summarization. However, such probabilistic topic models ignore word order and represent a given context as a bag of its words, thereby disregarding semantic information. 

To motivate our first task of extending probabilistic topic models to incorporate word order and language structure, assume that we conduct topic analysis on the following two sentences: 

{\small \texttt{Bear falls into market territory} and \texttt{Market falls into bear territory}} 

When estimating the probability of a word in a given context (here: $P(``bear" | \mbox{context})$), traditional topic models do not account for language structure since they ignore word order within the context and are based on ``bag-of-words" (BoWs) only. 
In this particular setting, the two sentences have the same unigram statistics, but are about different topics.
On deciding which topic generated the word ``bear" in the second sentence, the preceding words  ``market falls" make it more 
likely that it was generated by a topic that assigns a high probability to words related to {\it stock market trading}, where ``bear territory" is a colloquial expression in the domain. 
\enote{PG}{need advise from Hinrich about example of bear in the next line}
In addition, the language structure (e.g., syntax and semantics) is also ignored. 
For instance, the word ``bear" in the first sentence is a proper noun and subject while it is an object in the second. 
In practice, topic models also ignore functional words such as ``into", which may not be appropriate in some scenarios. 

Recently, \citet{PeterELMonaacl2018} have shown that a deep contextualized LSTM-based language model (LSTM-LM) is able to capture 
different language concepts in a layer-wise fashion, e.g., the lowest layer captures language syntax and topmost layer captures 
 semantics. However, in LSTM-LMs the probability of a word is a function 
of its sentence only and  word occurrences are modeled in a {\it fine granularity}.  Consequently, LSTM-LMs do not capture semantics at a document level.    
To this end, recent studies such as  {TDLM} \citep{Lau2017TopicallyDN}, Topic-RNN \citep{Dieng2016TopicRNNAR} and TCNLM \citep{Wang2018TopicCN} have integrated the merits of latent topic and neural language models (LMs);  
however, they have focused on improving LMs with global (semantics) dependencies using latent topics. 

Similarly, while bi-gram LDA based topic models \citep{wallach2006topic,wang2007topical} and n-gram based topic learning  \citep{HugoJMLR:82}
can capture word order in short contexts, they are unable to capture long term dependencies and language concepts.  In contrast, DocNADE variants \citep{Hugo:82, pankajgupta:2019iDocNADEe} learns word occurrences across documents 
i.e., {\it coarse granularity} (in the sense that the topic assigned to a given word occurrence equally depends on all the other words appearing in the same document); however since it is based on the BoW 
assumption all  language structure is ignored. In language modeling, \citet{mikolov2010recurrent} have shown that recurrent neural networks result in a significant reduction of perplexity over standard n-gram models.

\begin{figure*}[t]
  \centering
  \includegraphics[scale=0.7]{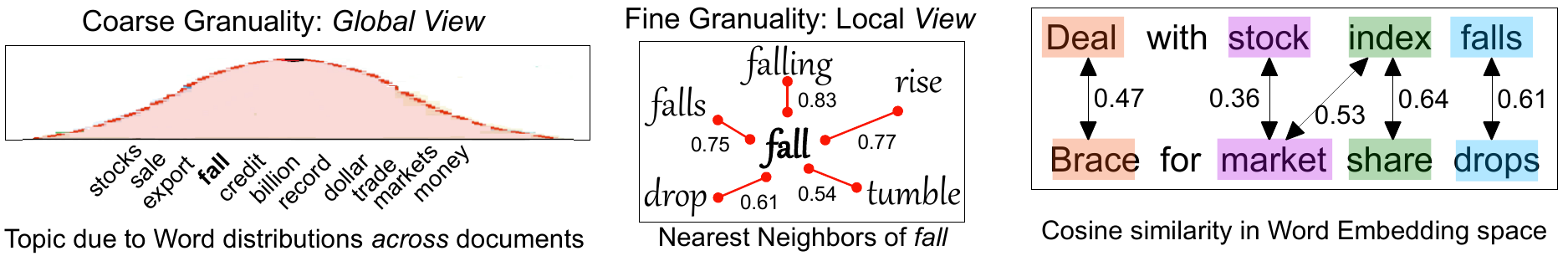}
\caption{(left): A topic-word distribution due to global exposure, obtained from the matrix ${\bf W}$ as row-vector.
(middle): Nearest neighbors in semantics space, represented by ${\bf W}$  in its column vectors. 
(right): BoW and cosine similarity illustration in distributed embedding space.
}
\label{fig:introexamples}
\end{figure*}

{\it Contribution 1}: We {\it introduce language structure}  into neural autoregressive topic models via a LSTM-LM, thereby accounting for word ordering (or semantic regularities), language concepts and long-range dependencies. 
This allows for the accurate prediction of words, 
where the probability of each word is a function of global and local (semantics) contexts, modeled via DocNADE and LSTM-LM, respectively.  
The proposed neural topic model is named as {\it contextualized}-{\it Document Neural Autoregressive Distribution Estimator} ({\it ctx-DocNADE}) 
and offers learning complementary semantics by combining  joint word and latent topic learning in a unified neural autoregressive framework. 
For instance, Figure \ref{fig:introexamples} (left and middle) shows the complementary topic and word semantics, based on TM and LM representations of the term ``fall". 
Observe that the topic captures the usage of ``fall" in the context of {\it stock market trading}, attributed to the global (semantic) view.  
 

While this is a powerful approach for incorporating language structure and word order 
in particular for long texts and corpora with many documents, 
learning from contextual information
remains challenging in settings with short texts
and few documents, since (1) limited word co-occurrences
or little context (2) significant word non-overlap in
such short texts and (3) small training corpus of documents lead to little evidence for learning word co-occurrences. 
However, distributional word representations
(i.e. word embeddings) \citep{pennington14glove2} have shown to capture both
the semantic and syntactic relatedness in words and demonstrated
impressive performance in NLP tasks.

For example, assume that we conduct
topic analysis over the two short text fragments: 
\texttt{Deal with stock index falls} and 
\texttt{Brace for market share drops}.  
Traditional topic models with ``BoW" assumption will not be able
to infer relatedness between word pairs such as ({\it falls}, {\it drops}) due to the lack of word-overlap and small context in the two phrases. 
However, in the distributed embedding space, the word pairs are semantically related as shown in Figure \ref{fig:introexamples} (left). 

Related work such as \citet{sahami2006web} employed web search results to improve the information in short texts and 
\citet{petterson2010word} introduced word similarity via thesauri and dictionaries into LDA. 
\citet{das2015gaussian} and \citet{nguyen2015gloveldaglovegmm} integrated word embeddings into LDA and Dirichlet Multinomial Mixture (DMM) \citep{nigam2000text} models. Recently, \cite{pankajgupta:2019iDocNADEe} extends DocNADE by introducing  pre-trained word embeddings in topic learning.  
However, they ignore the underlying language structure, e.g., word ordering, syntax, etc. 
In addition, DocNADE and its extensions outperform LDA and RSM topic models in terms of perplexity and IR. 

{\it Contribution 2}: We incorporate {\it distributed compositional priors} in DocNADE: we use  pre-trained word embeddings via LSTM-LM 
to supplement the multinomial topic model (i.e., DocNADE) in learning latent topic and textual representations  on a smaller corpus and/or short texts.   
Knowing similarities in a distributed space and integrating this complementary information via a LSTM-LM,  a topic representation is much more likely and coherent.  

Taken together, we combine the advantages of complementary learning and external knowledge, and couple topic- and language models with pre-trained word embeddings 
to model short and long text documents in a unified neural autoregressive framework, named as {\it ctx}-{\it DocNADEe}. 
Our approach learns better textual representations, which we quantify via generalizability (e.g., perplexity), interpretability (e.g., topic extraction and coherence) and applicability (e.g., IR and classification).



To illustrate our two {\it contributions},  we apply our modeling approaches to 7 long-text and 8 short-text datasets from diverse domains 
and demonstrate that our approach consistently outperforms state-of-the-art generative topic models. 
Our learned representations, result in  a gain of:  
(1) $4.6$\%  (.790 vs .755) in topic coherence,  
(2) $6.5$\%  (.615 vs .577) in precision at retrieval fraction 0.02, and 
(3) $4.4$\%  (.662 vs .634) in $F1$ for text classification, averaged over 6 long-text and 8 short-text datasets.

When applied to short-text and long-text documents, our proposed modeling approaches generate {\it con\underline{text}ualized \underline{to}pic \underline{vec}tors}, which we name  {\it textTOvec}. 
The {\it code} is available at \url{https://github.com/pgcool/textTOvec}. 

\begin{figure*}[t]
  \centering
  \includegraphics[scale=0.73]{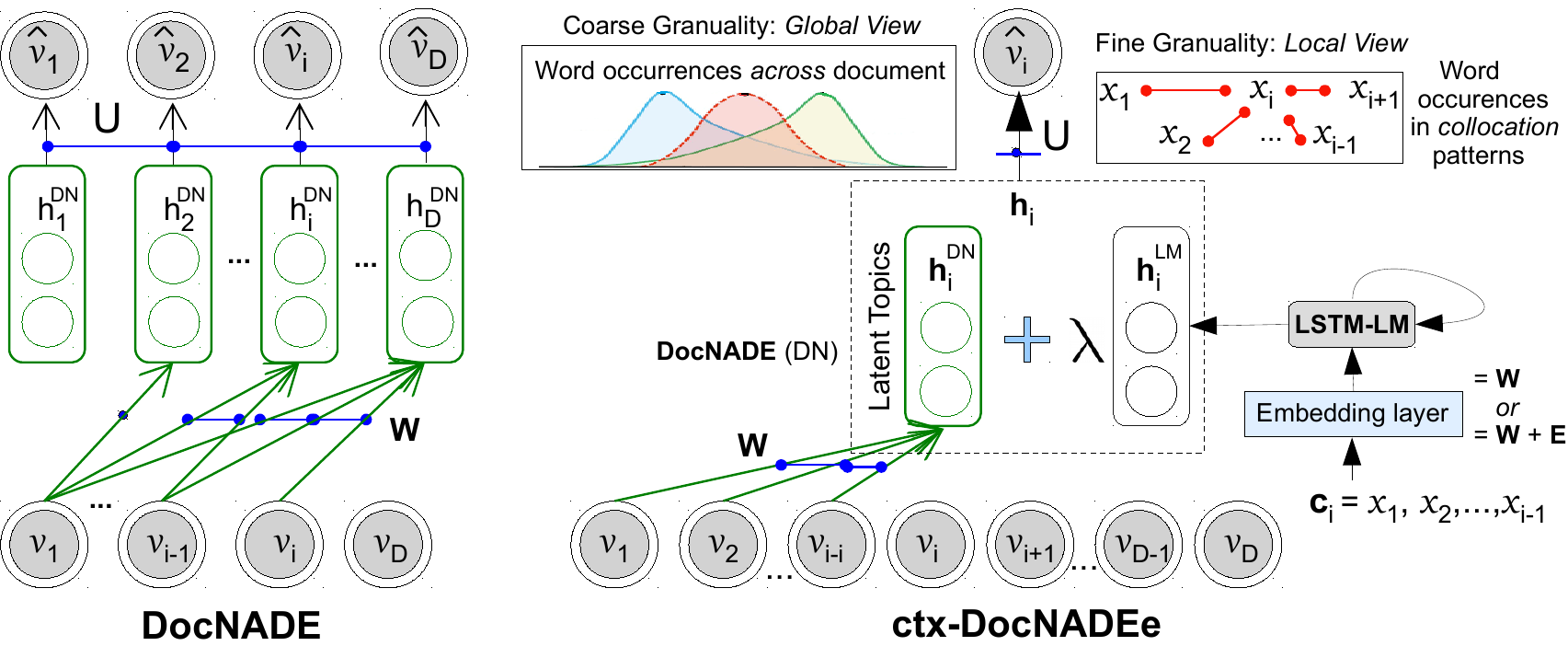}
\caption{(left): DocNADE for the document ${\bf v}$.     
(right): ctx-DocNADEe for the observable corresponding to $v_i \in {\bf v}$. 
Blue colored lines signify the connections that share parameters. The observations (double circle) for each word $v_i$ are multinomial, 
where $v_i$ is the index in the vocabulary of the $i$th word of the document. 
${\bf h}_i^{DN}$ and ${\bf h}_i^{LM}$ are hidden vectors from DocNADE and LSTM models, respectively for the target word $v_i$.  
Connections between each input $v_i$ and hidden units ${\bf h}_i^{DN}$ are shared. 
The symbol $\hat{v}_{i}$ represents the autoregressive conditionals $p(v_i | {\bf v}_{< i})$, computed using     
${\bf h}_i$ which is a weighted sum of ${\bf h}_i^{DN}$ and ${\bf h}_i^{LM}$ in ctx-DocNADEe.  
}
\label{fig:iDocNADELM}
\end{figure*}

\section{Neural Autoregressive Topic Models}

Generative models are based on estimating the probability distribution of multidimensional data, implicitly requiring modeling complex dependencies. 
Restricted Boltzmann Machine (RBM) \citep{Hinton06} and its variants \citep{larochelle2008classification} are probabilistic undirected models of binary data. RSM \citep{Salakhutdinov:82} and its variants \citep{Gupta:85} are generalization of the RBM, that are used to model word counts.  
However, estimating the complex probability distribution of the underlying high-dimensional observations is intractable.
To address this challenge, NADE \citep{larochelle2011neural} decomposes  the joint distribution of binary observations into autoregressive conditional distributions, 
each modeled using a feed-forward network. 
Unlike for RBM/RSM, this leads to tractable gradients of the data negative log-likelihood. 


\subsection{Document Neural Autoregressive Topic Model (DocNADE)}

An extension of NADE and RSM, DocNADE \citep{Hugo:82} models collections of documents as orderless bags of words (BoW approach), thereby disregarding any language structure. 
In other words, it is trained to learn word representations reflecting the underlying topics of the documents only, 
ignoring syntactical and semantic features as those encoded in word embeddings \citep{bengio2003neural,mikolov2013efficientword2vec,pennington14glove2,PeterELMonaacl2018}. 

DocNADE \citep{HugoJMLR:82} represents a document by transforming its BoWs into a sequence ${\bf v}=[v_1, ..., v_D]$ of size $D$, where 
each element $v_i \in \{1,2,..., K\}$ corresponds to a multinomial observation (representing a word from a vocabulary of size $K$). 
Thus, $v_i$ is the index in the vocabulary of the $i$th word of the document {\bf v}.  
DocNADE models the joint distribution $p({\bf v})$ of all words $v_i$ by decomposing it as  $p({\bf v})= \prod_{i=1}^{D} p(v_i | {\bf v}_{<i})$, where 
each autoregressive conditional  $p(v_i | {\bf v}_{<i})$  for the word observation $v_i$ is computed using the preceding observations  
${\bf v}_{<i} \in \{v_1, ...,v_{i-1}\}$ in a  feed-forward neural network for $i \in \{1,...D\}$,
{\small \begin{eqnarray}\label{docnadehidden}
{\bf h}_i^{DN}({\bf v}_{<i})  =  g ({\bf e} + \sum_{k<i} {\bf W}_{:, v_k}) \ \  \mbox{and} \ \ 
p (v_i = w | {\bf v}_{<i})  = \frac{\exp (b_w + {\bf U}_{w,:} {\bf h}_i^{DN}({\bf v}_{<i}))}{\sum_{w'} \exp (b_{w'} + {\bf U}_{w',:} {\bf h}_i^{DN} ({\bf v}_{<i}))}
\end{eqnarray}}
where $g(\cdot)$ is an activation function, 
${\bf U} \in \mathbb{R}^{K \times H}$ is a weight matrix connecting hidden to output,  
${\bf e} \in \mathbb{R}^H$ and ${\bf b} \in \mathbb{R}^K$ are bias  vectors,
${\bf W} \in \mathbb{R}^{H \times K}$ is a word representation matrix in which a column ${\bf W}_{:,v_i}$ is a vector representation of  the word $v_i$ in the vocabulary, and   
$H$ is the number of hidden units (topics). 
The log-likelihood of any document ${\bf v}$ of any arbitrary length is given by: 
$\mathcal{L}^{DN}({\bf v})  =  \sum_{i=1}^{D} \log p (v_i | {\bf v}_{<i})$. 
Note that the past word observations ${\bf v}_{<i}$ are orderless due to BoWs, and may not correspond to 
the  words preceding the $i$th word in the document itself.

\begin{minipage}{.36\linewidth}
  \begin{algorithm}[H]
\caption{{\small 
Computation of $\log p({\bf v})$}}\label{algo:computelogpv} 
\small
{ 
\begin{algorithmic}[1]
\Statex \textbf{Input}: A training document {\bf v}
\Statex \textbf{Input}: Word embedding matrix {\bf E}
\Statex \textbf{Output}: $\log p({\bf v})$
\State ${\bf a} \gets {\bf e}$ 
\State $ q({\bf v}) = 1$
\For{$i$ from $1$ to $D$}
        \State compute ${\bf h}_{i}$  and $p(v_{i} | {\bf v}_{<i})$
	\State $ q({\bf v}) \gets  q({\bf v}) p(v_{i} | {\bf v}_{<i})$
        \State ${\bf a} \gets {\bf a} + {\bf W}_{:, v_{i}}$  
\EndFor
\State $\log p({\bf v}) \gets  \log q({\bf v})$
\end{algorithmic}}
\end{algorithm}
\end{minipage}%
\hfill
\begin{minipage}{.60\linewidth}
\begin{table}[H]
\centering
\renewcommand*{\arraystretch}{1.5}
\resizebox{.98\linewidth}{!}{
\begin{tabular}{c|c|c}
     {\it model}                                           &                  ${\bf h}_{i}$                                          &              $p(v_{i} | {\bf v}_{<i})$              \\ \hline\hline
\multirow{2}{*}  {\texttt{DocNADE}}           &                 ${\bf h}_{i}^{DN}  \gets g({\bf a})$          &        \multirow{2}{*}{equation \ref{docnadehidden}}                                                  \\
                                                                &                 $ {\bf h}_{i}   \gets   {\bf h}_{i}^{DN} $    &                                          \\ \hline

\multirow{2}{*}  {\texttt{ctx-DocNADE}}      &                 ${\bf h}_{i}^{LM} \gets \mbox{LSTM}({\bf c}_i, \mbox{embedding}= {\bf W})$                              &                            \multirow{2}{*}{equation \ref{hctxdocnade}}                                      \\ 
                                                               &                  ${\bf h}_{i}  \gets   {\bf h}_{i}^{DN} + \lambda \  {\bf h}_{i}^{LM}$                              &                                                                \\ \hline

\multirow{2}{*}{\texttt{ctx-DocNADEe}}     &               ${\bf h}_{i}^{LM} \gets \mbox{LSTM}({\bf c}_i, \mbox{embedding}={\bf W} + {\bf E})$                                 &                          \multirow{2}{*}{equation \ref{hctxdocnade}}                                        \\
                                                               &               ${\bf h}_{i}  \gets   {\bf h}_{i}^{DN} + \lambda \  {\bf h}_{i}^{LM}$                                  &                                                               \\ 
\end{tabular}}
\caption{Computation of ${\bf h}_{i}$ and $p(v_{i} | {\bf v}_{<i})$ in {\texttt{DocNADE}}, 
{\texttt{ctx-DocNADE}}  and {\texttt{ctx-DocNADEe}} models, correspondingly used in estimating $\log p({\bf v})$ 
(Algorithm \ref{algo:computelogpv}).}
\label{tab:computelogpv}
\end{table}
\end{minipage}

\subsection{Deep Contextualized DocNADE with Distributional Semantics}\label{sec:ctx-docnade}
We propose two extensions of the DocNADE model: 
(1) {\it ctx-DocNADE}: introducing language structure via LSTM-LM  and
(2) {\it ctx-DocNADEe}: incorporating external knowledge via 
pre-trained word embeddings ${\bf E}$, to model short and long texts. 
The unified network(s) account for the ordering of words, 
syntactical and semantic structures in a language, long and short term dependencies, 
as well as external knowledge, thereby circumventing the major drawbacks of BoW-based 
representations.   
 
Similar to DocNADE, ctx-DocNADE models each document $\bf v$ as a sequence of multinomial observations. Let $[x_1, x_2, ..., x_N] $ be a sequence of N words in a given document, where $x_i$ is represented by an embedding vector of dimension, $dim$. 
Further, for each element $v_i \in {\bf v}$, let ${\bf c}_{i} = [x_1, x_2, ..., x_{i-1}] $ be the context (preceding words) of $i$th word in the document. 
Unlike in DocNADE, the conditional probability of the word $v_i$ in ctx-DocNADE (or ctx-DocNADEe) is a function of two hidden vectors:   
${\bf h}_i^{DN}({\bf v}_{<i})$ and ${\bf h}_i^{LM}({\bf c}_{i})$, stemming from the DocNADE-based and LSTM-based components of ctx-DocNADE, respectively: 
{\small 
\begin{eqnarray}\label{hctxdocnade}
{\bf h}_i({\bf v}_{<i})  =  {\bf h}_i^{DN}({\bf v}_{<i}) + \lambda \ \ {\bf h}_i^{LM}({\bf c}_{i}) 
\ \mbox{and} \ p (v_i = w | {\bf v}_{<i})  = \frac{\exp (b_w + {\bf U}_{w,:} {\bf h}_i ({\bf v}_{<i}))}{\sum_{w'} \exp (b_{w'} + {\bf U}_{w',:} {\bf h}_i ({\bf v}_{<i}))}
\end{eqnarray}}
where ${\bf h}_i^{DN}({\bf v}_{<i})$ is computed as in DocNADE (equation \ref{docnadehidden}) 
and $\lambda$ is  the mixture weight of the LM component, which can be optimized during training (e.g., based on the validation set). 
The second term ${\bf h}_i^{LM}$ is a context-dependent representation and output of an LSTM layer at position $i-1$
over input sequence ${\bf c}_{i}$, trained to predict the next word $v_i$.  The LSTM offers history for the $i$th word via modeling temporal dependencies in the input sequence, ${\bf c}_{i}$. 
The conditional distribution for each word $v_i$ is estimated by equation \ref{hctxdocnade}, where 
the unified network of DocNADE and LM combines global and context-dependent representations.   
Our model is jointly optimized to maximize the pseudo log likelihood, $\log p({\bf v}) \approx \sum_{i=1}^{D} \log p(v_i | {\bf v}_{<i})$ with stochastic gradient descent. See \cite{Hugo:82} for more details on training from bag of word counts.

In the weight matrix $\bf W$ of DocNADE \citep{Hugo:82},  
each row vector ${\bf W}_{j,:}$ encodes topic information for the $j$th hidden topic feature 
and 
each column vector ${\bf W}_{:, v_{i}}$ is a vector for the word $v_i$. To obtain complementary semantics, we exploit this property and expose
  $\bf W$ to both global and local influences by sharing $\bf W$  in the DocNADE and LSTM-LM componenents. 
Thus, the embedding layer of LSTM-LM component represents the column vectors. 

{\it ctx-DocNADE}, in this realization of the unified network the embedding layer in the LSTM component is randomly initialized. This extends DocNADE by accounting for the ordering of words
and language concepts via context-dependent representations for each word in the document.    

{\it ctx-DocNADEe}, the second version extends ctx-DocNADE with distributional priors, where the embedding layer in the LSTM component is initialized by the sum of a pre-trained embedding matrix ${\bf E}$ and the weight matrix ${\bf W}$. 
Note that ${\bf W}$ is a model parameter; however  ${\bf E}$ is a static prior.

Algorithm \ref{tab:computelogpv} and Table \ref{tab:computelogpv} show the 
 $\log p({\bf v})$ for a document ${\bf v}$ in three different settings: {\it DocNADE}, 
{\it ctx-DocNADE} and {\it ctx-DocNADEe}. In the DocNADE component, since the weights in the matrix ${\bf W}$ are tied, 
the linear activation ${\bf a}$ can be re-used in every hidden layer and computational complexity reduces to $O(HD)$, where H is the size of each hidden layer.
In every epoch, we run an LSTM over the sequence of words in the document and extract hidden vectors $h_i^{LM}$, 
corresponding to ${\bf c}_i$ for every target word $v_i$. Therefore, the computational complexity in  {ctx-DocNADE} or {ctx-DocNADEe} is $O(HD+\mathfrak{N})$, 
where  $\mathfrak{N}$ is the total number of edges in the LSTM network \citep{LSTMhochreiter1997long,sak2014long}.  
The trained models can be used to extract a {\it textTOvec} representation, i.e.,   
{\small ${\bf h}({\bf v}^{*})  =  {\bf h}^{DN}({\bf v}^{*}) + \lambda \ {\bf h}^{LM}({\bf c}^{*}_{N+1})$} for the text ${\bf v}^{*}$ of length ${\bf D}^{*}$, where
{\small ${\bf h}^{DN}({\bf v}^{*}) = g ({\bf e} + \sum_{k \le {\bf D}^{*}} {\bf W}_{:, v_k})$}  and 
{\small ${\bf h}^{LM}({\bf c}^{*}_{N+1}) = \mbox{LSTM}({\bf c}^{*}_{{N+1}},  \mbox{embedding} = {\bf W} \mbox{ or } ({\bf W}+{\bf E}))$}. 

\begin{table*}[t]
\center
\small
\renewcommand*{\arraystretch}{1.2}
\resizebox{.95\textwidth}{!}{
\setlength\tabcolsep{3.pt}
\begin{tabular}{r|rrrrrrrr||r|rrrrrrrr}
\hline
\multicolumn{9}{c||}{\texttt{short-text}} & \multicolumn{9}{c}{\texttt{long-text}} \\ 
 \multicolumn{1}{c|}{\bf Data} &  \multicolumn{1}{c}{\bf Train} &  \multicolumn{1}{c}{\bf Val} & \multicolumn{1}{c}{\bf Test} &  \multicolumn{1}{c}{$|${\bf RV}$|$}& \multicolumn{1}{c}{$|${\bf FV}$|$}  & \multicolumn{1}{c}{\bf L} & \multicolumn{1}{c}{\bf C} & \multicolumn{1}{c||}{\bf Domain}      
&\multicolumn{1}{c|}{\bf Data} &  \multicolumn{1}{c}{\bf Train} &  \multicolumn{1}{c}{\bf Val} & \multicolumn{1}{c}{\bf Test} &  \multicolumn{1}{c}{$|${\bf RV}$|$} &  \multicolumn{1}{c}{$|${\bf FV}$|$}  & \multicolumn{1}{c}{\bf L} & \multicolumn{1}{c}{\bf C} & \multicolumn{1}{c}{\bf Domain}   \\ \hline 
20NSshort             & 1.3k & 0.1k & 0.5k &    1.4k       &   1.4k    &  13.5     &   20      &     News         & 20NSsmall             & 0.4k &  0.2k & 0.2k  &      2k       &   4555   &   187.5     &    20      & News       \\
TREC6                  & 5.5k  &  0.5k &  0.5k  &      2k          &   2295 &   9.8          & 6        &   Q\&A      & Reuters8             & 5.0k &  0.5k & 2.2k  &      2k    &   7654   &   102       &    8      & News   \\
R21578title$^\dagger$ &  7.3k &  0.5k & 3.0k     &   2k        &   2721       &   7.3   &     90  &   News   & 20NS                  & 7.9k & 1.6k & 5.2k  &     2k       &   33770    &   107.5     &   20       &   News       \\
Subjectivity           &  8.0k &  .05k &  2.0k  & 2k       &   7965   &   23.1       &  2     &   Senti      & R21578$^\dagger$ &  7.3k&  0.5k & 3.0k  &      2k           &   11396    &   128       &   90      &    News   \\

Polarity                 & 8.5k &  .05k & 2.1k &    2k       &   7157    &   21.0     &   2              &  Senti        & BNC   &  15.0k &  1.0k & 1.0k  &      9.7k       &   41370  &  1189       &  -    &  News  \\ 

TMNtitle                & 22.8k  &  2.0k & 7.8k  &       2k      &   6240     &   4.9    &     7    &   News        & SiROBs$^\dagger$ &  27.0k &  1.0k & 10.5k  &      3k       &   9113  &  39       &  22    &  Indus    \\

TMN                 & 22.8k  &  2.0k & 7.8k  &      2k         &   12867  &    19          &       7   &  News             &       AGNews              & 118k & 2.0k &   7.6k &        5k       &   34071   &  38       &     4    &   News   \\

AGnewstitle           & 118k &  2.0k &  7.6k  &        5k         &   17125   &  6.8      &    4      &  News     &            &   &   &    &         &       &      &        &  
\end{tabular}}
\caption{Data statistics: Short/long texts and/or small/large corpora from diverse domains. 
Symbols-  Avg: average, $L$: avg text length (\#words), $|RV|$ and $|FV|$: size of reduced (RV) and full vocabulary (FV), 
$C$: number of classes, Senti: Sentiment, Indus: Industrial, `k':thousand and $\dagger$: multi-label.   
For short-text,  $L$$<$$25$.}
\label{datastatistics}
\end{table*}

{\it ctx-DeepDNEe}: DocNADE and LSTM can be extended to a deep, 
multiple hidden layer architecture by adding new hidden layers as in a regular deep
feed-forward neural network, allowing for improved performance.  
In the deep version, the first hidden layer is computed in 
an analogous fashion to DocNADE variants (equation \ref{docnadehidden} or \ref{hctxdocnade}). 
Subsequent hidden layers are computed as:
{\[
{\bf h}_{i, d}^{DN} ({\bf v}_{<i}) = g ({\bf e}_d + {\bf W}_{i,d} \cdot {\bf h}_{i,d-1}({\bf v}_{<i}))  
\ \ \mbox{or} \ \ 
{\bf h}_{i, d}^{LM} ({\bf c}_{i}) = deepLSTM({\bf c}_i,  \mbox{depth}=d)
\]}
for $d=2,... n$, where n is the total number of hidden layers (i.e., depth) in the
deep feed-forward and LSTM networks. 
For $d$=$1$, the hidden vectors ${\bf h}_{i, 1}^{DN}$ and ${\bf h}_{i, 1}^{LM}$ 
correspond to equations \ref{docnadehidden} and \ref{hctxdocnade}.
The conditional $p (v_i = w | {\bf v}_{<i})$ is computed using the last layer $n$, 
i.e., ${\bf h}_{i, n} = {\bf h}_{i, n}^{DN} + \lambda \ {\bf h}_{i, n}^{LM}$.  

\begin{table*}[t]
\center
\small
\renewcommand*{\arraystretch}{1.2}
\resizebox{.95\textwidth}{!}{
\begin{tabular}{r|rr|rr|rr|rr|rr|rr|rr|rr||rr}

 \multirow{2}{*}{\bf Model} 
&  \multicolumn{2}{c|}{\bf 20NSshort}      &  \multicolumn{2}{c|}{\bf TREC6}    &  \multicolumn{2}{c|}{\bf R21578title}    &  \multicolumn{2}{c|}{\bf Subjectivity}     
&  \multicolumn{2}{c|}{\bf Polarity}  &  \multicolumn{2}{c|}{\bf TMNtitle}    &  \multicolumn{2}{c|}{\bf TMN}   &  \multicolumn{2}{c||}{\bf AGnewstitle}   &  \multicolumn{2}{c}{\bf Avg} \\ \cline{2-19}

                       &    IR      &  $F1$           &    IR      &  $F1$         &    IR      &  $F1$             &    IR      &  $F1$             
                       &    IR      &  $F1$         &    IR      &  $F1$        &    IR      &  $F1$       &    IR      &  $F1$           &    IR      &  $F1$          \\ \hline

{\it glove}(RV)        & .236      & \underline{.493}       & .480     &  .798     & .587      &  \underline{.356}     & .754     & .882              
                             & .543     &  .715         & .513      &  .693     & .638     & .736       & .588     & 814      &  .542       & .685     \\

{\it glove}(FV)        & .236      &  .488        &  .480     & .785       & .595       & .356           & .775    & .901         
                             & .553     &  .728        &  .545     & \underline{.736}        & .643         & \underline{.813}         &  .612    &  \underline{.830}     &  .554        &  .704        \\

{\it doc2vec}          & .090      &  .413       & .260     &  .400     & .518      &  .176     & .571     & .763              
                             & .510     &  .624         & .190      &  .582     & .220     & .720       & .265     & .600      &  .328       & .534     \\

{\it Gauss-LDA}        & .080      &  .118       & .325     &  .202     & .367      &  .012     & .558     & .676              
                             & .505     &  .511         & .408      &  .472     & .713     & .692       & .516     & .752      &  .434       & .429     \\

{\it glove-DMM}        & .183      &  .213       &  .370     & .454        & .273         & .011        & .738    & .834          
                               & .515     &  .585        &  .445     & .590       &  .551       &  .666        &  .540     & .652        &  .451       & .500        \\

{\it glove-LDA}        &  .160      &  .320           &  .300     & .600       & .387         & .052          & .610      &  .805          
                            & .517     &  .607                &  .260     & .412        & .428        & .627         &     .547    &   .687       &  .401       & .513              \\

{\it TDLM}        & .219      &  .308       & .521     &  .671     & .563      &  .174     & .839     & .885              
                             & .520     &  .599         & .535      &  .657     & .672     & .767       & .534     & .722      &  .550       & .586     \\

{\it DocNADE(RV)}       & .290      &  .440       & .550     &  .804     & .657      &  .313     & .820     & .889              
                             & .560     &  .699         & .524      &  .664     & .652     & .759       & .656     & .819      &  .588       & .673     \\

{\it DocNADE(FV)}        & .290      &  .440       & .546     &  .791     & .654      &  .302     & .848     & .907              
                             & .576     &  .724         & .525      &  .688     & .687     & .796       & .678     & .821      &  .600       & .683     \\


{\it DeepDNE}        & .100      &  .080       & .479     &  .629     & .630      &  .221     & .865     & .909              
                             & .503     &  .531         & .536      &  .661     & .671     & .783       & .682     & .825      &  .558       & .560     \\
\hline

{\it ctx-DocNADE}       & .296      &  .440       & .595     &  .817     & .641      &  .300     & .874     & .910              
                             & .591     &  .725         & .560      &  .687     & .692     & .793       & .691     & .826      &  .617       & .688     \\

{\it ctx-DocNADEe}        & {\bf .306}      &  .490       & .599    &  \underline{.824}     & {\bf .656}      &  .308     & .874     & .917              
                             & {\bf .605}     &  \underline{.740}         & {\bf .595}      &  .726     & {\bf .698}     & .806       & {\bf .703}     & .828      &  {\bf .630}       & \underline{.705}     \\

{\it ctx-DeepDNEe}        & .278      &  .416       & {\bf .606}     &  .804     & .647      &  .244     & {\bf .878}     & \underline{.920}              
                             & .591     &  .723         & .576      &  .694     & .687     & .796       & .689     & .826      &  .620       & .688     \\

\end{tabular}}
\caption{State-of-the-art comparison: IR (i.e, IR-precision at 0.02 fraction) and classification $F1$ for {\it short} texts, where $Avg$: average over the row values, 
the {\bf bold} and \underline{underline}: the maximum for IR and F1, respectively.}
\label{PPLIRF1scoresshorttext}
\end{table*}

\section{Evaluation}

We apply our modeling approaches (in improving topic models, i.e, DocNADE using language concepts from LSTM-LM) to 8 short-text and 7 long-text
datasets of varying size with single/multi-class labeled documents
from public as well as industrial corpora. 
We present four quantitative measures in evaluating topic models: generalization (perplexity), topic coherence, text retrieval and categorization. 
See the {\it appendices}
for the data description and example texts. Table \ref{datastatistics}
shows the data statistics, where 20NS and R21578 signify 20NewsGroups and Reuters21578, respectively.

{\it Baselines}: While, we evaluate our multi-fold contributions on four tasks: generalization (perplexity), topic coherence, text retrieval and categorization, we compare performance of our proposed models \texttt{ctx-DocNADE} and \texttt{ctx-DocNADEe} with related baselines based on: 
{\bf (1)} word representation: \texttt{glove} \citep{pennington14glove2}, where a document is represented by summing the embedding vectors of it's words,  
{\bf (2)} document  representation: \texttt{doc2vec} \citep{le14sentences}, 
{\bf (3)} LDA based BoW TMs: \texttt{ProdLDA} \citep{SrivastavaProdLDAICLR2017} and \texttt{SCHOLAR}\footnote{focuses on incorporating meta-data (author, date, etc.) into TMs; SCHOLAR w/o meta-data $\equiv$ ProdLDA}  \citep{card2017neural}
{\bf (4)} neural BoW TMs: \texttt{DocNADE} 
and \texttt{NTM} \citep{cao2015novel} and ,    
{\bf (5)} TMs,  including pre-trained word embeddings: \texttt{Gauss-LDA} (\texttt{GaussianLDA}) \citep{das2015gaussian}, and 
\texttt{glove-DMM}, \texttt{glove-LDA} \citep{nguyen2015gloveldaglovegmm}. 
{\bf (6)} jointly\footnote{though focused on improving language models using topic models, different to our motivation} trained topic and language models:  \texttt{TDLM} \citep{Lau2017TopicallyDN}, \texttt{Topic-RNN} \citep{Dieng2016TopicRNNAR} and \texttt{TCNLM} \citep{Wang2018TopicCN}.

\begin{minipage}{0.66\linewidth}
\begin{table}[H]
\center
\small
\renewcommand*{\arraystretch}{1.2}
\resizebox{.96\textwidth}{!}{
\setlength\tabcolsep{3.5pt}
\begin{tabular}{r|rr|rr|rr|rr|rr|rr||rr}

 \multirow{2}{*}{\bf Model} 
&  \multicolumn{2}{c|}{\bf 20NSsmall}      &  \multicolumn{2}{c|}{\bf Reuters8}    &  \multicolumn{2}{c|}{\bf 20NS}    &  \multicolumn{2}{c|}{\bf R21578}     
&  \multicolumn{2}{c|}{\bf SiROBs}    &  \multicolumn{2}{c||}{\bf AGnews}  &  \multicolumn{2}{c}{\bf Avg}    \\ \cline{2-15}


                            &    IR      &  $F1$        &    IR      &  $F1$        &    IR      &  $F1$       &    IR      &  $F1$     &    IR      &  $F1$     &    IR      &  $F1$      &    IR      &  $F1$         \\ \hline

{\it glove}(RV)        & .214      &  .442        &  .845     & .830         & .200       & .608         & .644      &  .316        & .273     &  .202       & .725        &  .870        & .483     & .544                 \\

{\it glove}(FV)        & .238      &  .494         &  .837     & .880        & .253    & .632           & .659    & .340        & .285     &  .217      &  .737      & .890       & .501     & .575                     \\

{\it doc2vec}          & .200      &  .450         & .586        &  .852       & .216         &  .691        & .524       & .215       & .282     & .226         & .387        &  .713       & .365       &  .524                  \\

{\it Gauss-LDA}         &  .090    &  .080         & .712        &  .557       & .142         &  .340        & .539       & .114       & .232     & .070         & .456        &  .818       & .361       &  .329                 \\

{\it glove-DMM}       & .060      &  .134         & .623        &  .453       & .092         &  .187        & .501       & .023       & .226     & .050         & -        &  -       & -       &  -                \\


{\it DocNADE(RV)}        & .270      &  .530         & {\bf .884}        &  .890       & .366        &  .644      &   {\bf .723}         &   .336        & .374       & .298       & .787     & .882              & .567       &  .596               \\

{\it DocNADE(FV)}         & .299      &  .509         & .879        &  \underline{ .907}       & .427        &  .727     & .715         &  \underline{ .340}        & .382       & .308       & .794     & .888             & .582       &  .613               \\ \hline

{\it ctx-DocNADE}        & .313      & \underline{ .526}         & .880        &  .898       & .472         &  .732          & .714       & .315       & .386     & .309         & .791        &  .890       & .592       &  .611         \\

{\it ctx-DocNADEe}         & {\bf .327}      &  .524         & .883        &  .900       & {\bf .486}         &  \underline{ .745}        & .721       & .332       & {\bf .390}     & \underline{ .311}         & {\bf .796}        &  \underline{ .894}       & {\bf .601}       &  \underline{ .618}          

\end{tabular}}
\caption{IR-precision at fraction 0.02 and classification $F1$ for {\it long} texts}
\label{PPLIRF1scoreslongtext}
\end{table}
\end{minipage}\hfill
\begin{minipage}{0.34\linewidth}
\begin{table}[H]
\center
\small
\renewcommand*{\arraystretch}{1.4}
\resizebox{1.0\textwidth}{!}{
\setlength\tabcolsep{3pt}
\begin{tabular}{lr|r||lr|r}
     &  \multicolumn{1}{c|}{\bf Model}      &  \multicolumn{1}{c||}{\bf PPL}  &       &  \multicolumn{1}{c|}{\bf Model}      &  \multicolumn{1}{c}{\bf PPL}  \\  \hline

\multirow{4}{*}{\rotatebox{90}{Subjectivity}}        &        \texttt{DocNADE}            &  980                           & \multirow{4}{*}{\rotatebox{90}{AGnewstitle}}        &        \texttt{DocNADE}            &  846  \\
                                                                   &        \texttt{ctx-DocNADE}      &  968           &                                         &        \texttt{ctx-DocNADE}      &  822   \\ 
                                                                   &        \texttt{ctx-DocNADEe}      &  966          &                                         &        \texttt{ctx-DocNADEe}      &  820   \\  \cline{2-3} \cline{5-6}

\multirow{4}{*}{\rotatebox{90}{Reuters8}}        &        \texttt{DocNADE}            &  283                          & \multirow{4}{*}{\rotatebox{90}{20NS}}        &        \texttt{DocNADE}            &  1375  \\
                                                                   &        \texttt{ctx-DocNADE}      &  276           &                                         &        \texttt{ctx-DocNADE}      &  1358   \\ 
                                                                   &        \texttt{ctx-DocNADEe}      &  272          &                                         &        \texttt{ctx-DocNADEe}      &  1361   \\   \cline{2-3} \cline{5-6}

\multirow{4}{*}{\rotatebox{90}{TMNtitle}}        &        \texttt{DocNADE}            &  1437                           & \multirow{4}{*}{\rotatebox{90}{20NSshort}}        &        \texttt{DocNADE}            &  646  \\
                                                                   &        \texttt{ctx-DocNADE}      &  1430           &                                         &        \texttt{ctx-DocNADE}      &  656   \\ 
                                                                   &        \texttt{ctx-DocNADEe}      &  1427          &                                         &        \texttt{ctx-DocNADEe}      &  648   
               
\end{tabular}}
\caption{Generalization: PPL}
\label{PPLscores}
\end{table}
\end{minipage}

{\it Experimental  Setup}: DocNADE is often trained on a reduced vocabulary (RV) after pre-processing (e.g., ignoring functional words, etc.); however, we also investigate training it on full text/vocabulary (FV)  (Table \ref{datastatistics}) 
and compute document representations to perform different evaluation tasks. The FV setting preserves the language structure, required by LSTM-LM, and allows a fair comparison of DocNADE+FV and ctx-DocNADE 
variants. 
We use the glove embedding of 200 dimensions. 
All the baselines and proposed models ({\it ctx-DocNADE}, {\it ctx-DocNADEe} and {\it ctx-DeepDNEe}) were run in the FV setting over 200 topics  to 
quantify the quality of the learned representations. To better initialize the complementary learning in ctx-DocNADEs, we perform a pre-training for 10 epochs with $\lambda$ set to $0$.   
See the {\it appendices} for the experimental setup and hyperparameters 
for the following tasks, including the ablation over $\lambda$ on validation set. 

We run TDLM\footnote{https://github.com/jhlau/topically-driven-language-model} \citep{Lau2017TopicallyDN} for all the short-text datasets to evaluate the quality of representations learned in the spare data setting. 
For a fair comparison, we set 200 topics and hidden size, and initialize with the same pre-trained word embeddings (i.e., glove) as used in the ctx-DocNADEe.

\begin{table*}[t]
\center
\small
\renewcommand*{\arraystretch}{1.2}
\resizebox{.97\textwidth}{!}{
\setlength\tabcolsep{3.pt}
\begin{tabular}{r|rr|rr|rr|rr|rr||r|rr|rr|rr|rr}

 \multirow{2}{*}{\bf Data}    &  \multicolumn{2}{c|}{\bf glove-DMM}    &  \multicolumn{2}{c|}{\bf glove-LDA}    &  \multicolumn{2}{c|}{\bf DocNADE}     
&  \multicolumn{2}{c|}{\bf ctx-DNE}  &  \multicolumn{2}{c||}{\bf ctx-DNEe}   
& \multirow{2}{*}{\bf Data}      &  \multicolumn{2}{c|}{\bf glove-DMM}    
&  \multicolumn{2}{c|}{\bf DocNADE}     
&  \multicolumn{2}{c|}{\bf ctx-DNE}  &  \multicolumn{2}{c}{\bf ctx-DNEe}  \\ 
                             &    W10      &  W20      &    W10      &  W20      &    W10      &  W20      &    W10      &  W20   &    W10      &  W20  &  
                             &    W10      &  W20       &    W10      &  W20      &    W10      &  W20      &    W10      &  W20       \\ \hline

{\it 20NSshort}       & .512          &  .575      &  .616         &  .767     & .669           & .779       & .682         & .794    &   {\bf .696}      &  {\bf .801}   
& {\it Subjectivity}   & .538          &  .433         & .613           & .749       & .629         & .767    &   {\bf .634}      &  {\bf .771}   \\

{\it TREC6}            & .410          &  .475      &  .551         &  .736     & .699           & .818       & {\bf .714}         & {\bf  .810}    &   .713      &  .809    
& {\it AGnewstitle}    & .584          &  .678      & .731           & .757       & .739         & .858    &  {\bf  .746}      &  {\bf .865}    \\

{\it R21578title}     & .364          &  .458      &  .478         &  .677     & .701           & .812       & .713         & .802    &   {\bf .723}      &  {\bf .834}    
&   {\it 20NSsmall}       & {\bf .578}         &  .548         &  .508         &  .628     & .546           & .667       & .565         & {\bf .692}    \\

{\it Polarity}         & .637          &  .363      &  .375         &  .468     & .610           & .742       & .611         & .756    &   {\bf .650}      &  {\bf .779}     
& {\it Reuters8}   & .372         &  .302           &  .583         &  .710     & .584           & .710       & {\bf .592}         & {\bf .714}    \\

{\it TMNtitle}        & .633          &  .778      &  .651         &  .798     & .712           & .822       & .716         & .831    &   {\bf .735}      &  {\bf .845}     
&  {\it 20NS}        & .458         &  .374           &  .606         &  .729     & .615           & .746       & {\bf .631}         & {\bf .759}     \\
\cline{12-20} \cline{12-20}
           
{\it TMN}            & .705          &  .444      &  .550         &  .683     & .642           & .762       & .639         & .759    &   {\bf .709}      &  {\bf .825}      
 &  { Avg (all)}    & .527         &  .452     &  .643         &  .755     & .654          & .772       & {\bf  .672}         &  {\bf .790} 
    
\end{tabular}}
\caption{Average coherence for {\it short} and {\it long} texts over 200 topics in FV setting, where {\it DocNADE} $\leftrightarrow$ {\it DNE}}
\label{coherencescores}
\end{table*}

\subsection{Generalization: Perplexity (PPL)}\label{sec:ppl}
To evaluate the generative performance of the topic models, we estimate the log-probabilities for the test documents and compute 
the average held-out perplexity ($PPL$) per word as, {\small $PPL = \exp \big( - \frac{1}{z} \sum_{t=1}^{z} \frac{1}{|{\bf v}^t|} \log p({\bf v}^{t}) \big)$},  
where $z$  and $|{\bf v}^t|$ are the total number of documents and words in a document ${\bf v}^{t}$.   
For DocNADE, the log-probability $\log p({\bf v}^{t}) \big)$ is computed using $\mathcal{L}^{DN}({\bf v})$; 
however, we ignore the mixture coefficient, i.e., $\lambda$=0 (equation \ref{hctxdocnade}) 
to compute the exact log-likelihood in  ctx-DocNADE versions. The optimal $\lambda$ is determined based on the validation set. 
Table \ref{PPLscores}  quantitatively shows the PPL scores, where the complementary learning with $\lambda=0.01$ (optimal) in ctx-DocNADE 
achieves lower perplexity than the baseline DocNADE for both short and long texts, 
e.g., ($822$ vs $846$) and ($1358$ vs $1375$) on {\it AGnewstitle} and {\it 20NS} \footnote{PPL scores in (RV/FV) settings: DocNADE (665/1375) outperforms ProdLDA (1168/2097) on 200 topics}  datasets, respectively in the FV setting. 

\subsection{Interpretability: Topic Coherence}\label{sec:topiccoherence}
We compute topic coherence 
\citep{Chang:82,Newman:82,Gupta:85} 
to assess the meaningfulness of the 
underlying topics captured. 
We choose the coherence measure proposed by \citet{Michael:82}   
, which identifies context features for each topic word using a sliding 
window over the reference corpus. Higher scores imply more coherent topics.

We use the gensim module 
({\small \emph{radimrehurek.com/gensim/models/coherencemodel.html}, {\it coherence type} = $c\_v$}) to estimate coherence 
for each of the 200 topics (top 10 and 20 words). 
Table \ref{coherencescores} shows average coherence over 200 topics, 
where the higher scores in ctx-DocNADE compared to DocNADE ($.772$ vs $.755$) suggest that the contextual information and language structure help 
in generating more coherent topics.    
The introduction of embeddings in ctx-DocNADEe boosts the topic coherence, 
leading to a gain of $4.6$\%  (.790 vs .755) on average over 11 datasets. 
Note that the proposed models also outperform the baselines methods glove-DMM and glove-LDA. 
Qualitatively, Table \ref{topiccoherenceexamples} illustrates an example topic from the 20NSshort text dataset 
for DocNADE,  ctx-DocNADE  and ctx-DocNADEe, where the inclusion of embeddings results in a more coherent topic.

\begin{table*}[t]
\center
\small
\renewcommand*{\arraystretch}{1.2}
\resizebox{\textwidth}{!}{
\begin{tabular}{l|ccc}
\multirow{2}{*} {\bf Model} &  \multicolumn{3}{c}{\bf Coherence (NPMI)}\\ 
 & 50 & 100  &  150 \\   \hline

 \multicolumn{4}{c}{({\it sliding window}=20)}   \\ 

LDA\# 				       &     .106    &    .119     &    .119  \\
NTM\# 					&    .081    &    .070    &     .072  \\
TDLM(s)\# 				&    .102    &    .106	&      .100  \\
TDLM(l)\#    				&    .095	 &    .101    &    	.104  \\
Topic-RNN(s)\#                        &     .102    &    .108    &     .102  \\
Topic-RNN(l)\#                           &   .100   &     .105   &      .097  \\
TCNLM(s)\#                               &    .114   &     .111   &      .107   \\
TCNLM(l)\#                                &     .101  &    .104   &      .102   \\ \hline
                         
DocNADE                                    &   .097     &   .095    &   .097   \\  \hdashline 
ctx-DocNADE*($\lambda$=0.2)         &   .102    &   .103     &   .102   \\
ctx-DocNADE*($\lambda$=0.8)        &    .106    &   .105     &   .104   \\
ctx-DocNADEe*($\lambda$=0.2)      &   .098    &   .101      &    -        \\
ctx-DocNADEe*($\lambda$=0.8)      &   .105   &    .104     &     -       \\  \hline

 \multicolumn{4}{c}{({\it sliding window}=110)}   \\

DocNADE                                  &   .133    &   .131     &   .132     \\ \hdashline
ctx-DocNADE*($\lambda$=0.2)      &   .134     &   .141     &   .138       \\
ctx-DocNADE*($\lambda$=0.8)      &   .139    &   .142     &   .140   \\
ctx-DocNADEe*($\lambda$=0.2)    &   .133    &   .139     &     -       \\
ctx-DocNADEe*($\lambda$=0.8)    &   .135    &   .141     &     -     

\end{tabular}
\qquad

\begin{tabular}{l|c|c}
{\bf Topic}   &  {\bf Model}    &   {\bf Topic-words (ranked by their probabilities in topic)} \\ \hline 

				&   TCNLM\#	   &    pollution, emissions, nuclear, waste, environmental \\ 
{\it environment}	&    ctx-DocNADE*  &    ozone, pollution, emissions, warming, waste   \\ 
                                & ctx-DocNADEe*    &    pollution, emissions, dioxide, warming, environmental \\ \hline

                               &    TCNLM\#              &       elections, economic, minister, political, democratic  \\ 
{\it politics}               &    ctx-DocNADE*       &    elections, democracy, votes, democratic, communist   \\ 
                               &    ctx-DocNADEe*     &    democrat, candidate, voters, democrats, poll    \\  \hline

                              &    TCNLM\#              &        album, band, guitar, music, film    \\ 
{\it art}                    &    ctx-DocNADE*         &   guitar, album, band, bass, tone      \\ 
                             &    ctx-DocNADEe*     &     guitar, album, pop, guitars, song\\ \hline 

                            &     TCNLM\#               &        bedrooms, hotel, garden, situated, rooms     \\ 
{\it facilities}         &    ctx-DocNADE*         &    bedrooms, queen, hotel, situated, furnished     \\ 
                            &     ctx-DocNADEe*      &    hotel, bedrooms, golf, resorts, relax     \\  \hline 

                            &    TCNLM\#                &       corp, turnover, unix, net, profits     \\ 
{\it business}        &    ctx-DocNADE*         &    shares, dividend, shareholders, stock, profits     \\ 
                            &    ctx-DocNADEe*      &     profits, growing, net, earnings, turnover      \\  \hline

                            &    TCNLM\#                &        eye, looked, hair, lips, stared      \\ 
{\it expression}    &   ctx-DocNADE*         &     nodded, shook, looked, smiled, stared      \\ 
                           &    ctx-DocNADEe*      &      charming, smiled, nodded, dressed, eyes    \\ \hline

                            &    TCNLM\#                &        courses, training, students, medau, education      \\ 
{\it education}       &   ctx-DocNADE*         &     teachers, curriculum, workshops, learning, medau      \\ 
                           &    ctx-DocNADEe*      &      medau, pupils, teachers, schools, curriculum    
\end{tabular}

}
\caption{(Left): Topic coherence (NMPI) scores of different models for 50, 100 and 150 topics on BNC dataset. 
The {\it sliding window} is one of the hyper-parameters for computing topic coherence \citep{Michael:82, Wang2018TopicCN}. 
A {\it sliding window} of 20 is used in TCNLM; in addition we also present results for a window of size 110.  
$\lambda$ is the mixture weight of the LM component in the topic modeling process, and  
(s) and (l) indicate small and large model, respectively.  
The symbol '-' indicates no result, since word embeddings of 150 dimensions are not available from glove vectors.
(Right): The top 5 words of seven learnt topics from our models and TCNLM. 
The asterisk (*) indicates our proposed models and (\#) taken from TCNLM \citep{Wang2018TopicCN}. }
\label{topiccoherenceaddtionalbaselines}
\end{table*}

{\it Additional Baslines}:
We further compare our proposed models to other approaches that combining topic and language models, 
such as TDLM \citep{Lau2017TopicallyDN}, Topic-RNN \citep{Dieng2016TopicRNNAR} and TCNLM \citep{Wang2018TopicCN}. However, the related studies focus on improving language models  using topic models: in contrast, the focus of our work is on improving topic models for textual representations (short-text or long-text documents) by incorporating language concepts (e.g., word ordering, syntax, semantics, etc.)  and external knowledge (e.g., word embeddings) via neural language models, as discussed in section \ref{introduction}. 

To this end, we follow the experimental setup of the most recent work, TCNLM and {\it quantitatively} compare the performance of our models (i.e., ctx-DocNADE and ctx-DocNADEe) in terms of topic coherence (NPMI) on BNC dataset. Table \ref{topiccoherenceaddtionalbaselines} (left) shows NPMI scores of different models, where   
the results suggest that our contribution (i.e., ctx-DocNADE) of introducing language concepts into BoW topic model (i.e., DocNADE) improves topic coherence\footnote{NPMI over (50/200) topics learned on 20NS by: ProdLDA (.24/.19) and 
DocNADE (.15/.12) in RV setting}. 
The better performance for high values of $\lambda$ illustrates the relevance of the LM component for topic coherence (DocNADE corresponds to  $\lambda$=0). Similarly, the inclusion of word embeddings (i.e., ctx-DocNADEe) results in more coherent topics than the baseline DocNADE. 
Importantly, while ctx-DocNADEe is motivated by sparse data settings, the BNC dataset is neither a collection of short-text nor a corpus of few documents. Consequently, ctx-DocNADEe does not show improvements in topic coherence over ctx-DocNADE.

In Table \ref{topiccoherenceaddtionalbaselines} (right), we further {\it qualitatively} show the top 5 words of seven topics (topic name summarized by \cite{Wang2018TopicCN}) from TCNML and our models.  
Observe that ctx-DocNADE captures a topic {\it expression} that is a collection of only verbs in the past participle. 
Since the BNC dataset is unlabeled, we are here restricted to comparing model performance in terms of topic coherence only.

\makeatletter
\def\labelonly{BDF}
\def\labelcheck#1{
    \edef\pgfmarshal{\noexpand\pgfutil@in@{#1}{\labelonly}}
    \pgfmarshal
    \ifpgfutil@in@[#1]\fi
}
\makeatother

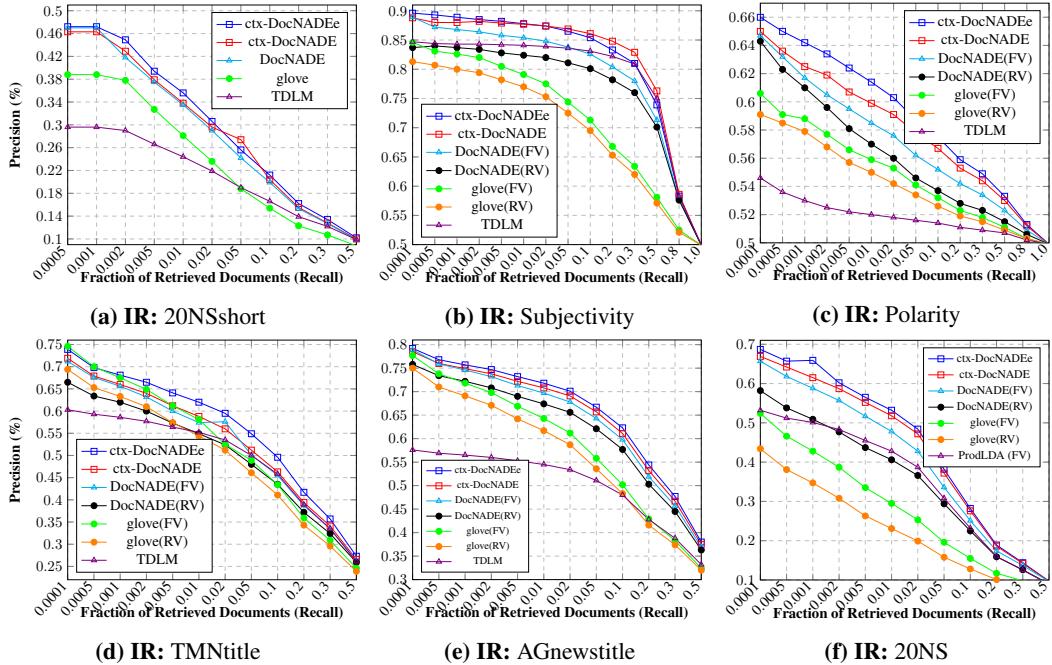
\begin{figure*}[t]
\centering

\begin{subfigure}{0.33\textwidth}
\centering
\begin{tikzpicture}[scale=0.56][baseline]
\begin{axis}[
    xlabel={\bf Fraction of Retrieved Documents (Recall)},
    ylabel={\bf Precision (\%)},
    xmin=0, xmax=10,
    ymin=0.09, ymax=0.51,
   /pgfplots/ytick={.10,.14,...,.51},
    xtick={0,1,2,3,4,5,6,7,8,9, 10,11,12},
    xticklabels={0.0005, 0.001, 0.002, 0.005, 0.01, 0.02, 0.05, 0.1, 0.2, 0.3, 0.5},
    x tick label style={rotate=45,anchor=east},
    legend pos=north east,
    ymajorgrids=true, xmajorgrids=true,
    grid style=dashed,
]

\addplot[
	color=blue,
	mark=square,
	]
	plot coordinates {
    (0, 0.472)
    (1, 0.472)
    (2, 0.449)
    (3, 0.394)
    (4, 0.356)
    (5, 0.306)
    (6, 0.256)
    (7, 0.212)
    (8, 0.162)
    (9, 0.134)
    (10, 0.102)
    (11, 0.077)
    (12, 0.066)
	};
\addlegendentry{ctx-DocNADEe}

\addplot[
	color=red,
	mark=square,
	]
	plot coordinates {
    (0, 0.463)
    (1, 0.463)
    (2, 0.429)
    (3, 0.379)
    (4, 0.338)
    (5, 0.296)
    (6, 0.274)
    (7, 0.204)
    (8, 0.155)
    (9, 0.128)
    (10, 0.10)
    (11, 0.076)
    (12, 0.066)
	};
\addlegendentry{ctx-DocNADE}

\addplot[
	color=cyan,
	mark=triangle,
	]
	plot coordinates {
    (0, 0.470)
    (1, 0.470)
    (2, 0.418)
    (3, 0.375)
    (4, 0.335)
    (5, 0.290)
    (6, 0.242)
    (7, 0.199)
    (8, 0.153)
    (9, 0.127)
    (10, 0.099)
    (11, 0.076)
    (12, 0.066)
    
	};
\addlegendentry{DocNADE}

\addplot[
	color=green,
	mark=*,
	]
	plot coordinates {
    (0, 0.388)
    (1, 0.388)
    (2, 0.378)
    (3, 0.327)
    (4, 0.281)
    (5, 0.236)
    (6, 0.188)
    (7, 0.154)
    (8, 0.123)
    (9, 0.107)
    (10, 0.088)
    (11, 0.072)
    (12, 0.066)
	};
\addlegendentry{glove}

\addplot[
	color=violet,
	mark=triangle,
	]
	plot coordinates {
    (0, 0.296)
    (1, 0.296)
    (2, 0.290)
    (3, 0.266)
    (4, 0.244)
    (5, 0.219)
    (6, 0.190)
    (7, 0.166)
    (8, 0.139)
    (9, 0.122)
    (10, 0.099)
    (11, 0.076)
    (12, 0.065)
	};
\addlegendentry{TDLM}

\end{axis}
\end{tikzpicture}%
\caption{{\bf IR:} 20NSshort} \label{IR20NSshort}
\end{subfigure}\hspace*{\fill}%
~~~%
\begin{subfigure}{0.33\textwidth}
\centering
\begin{tikzpicture}[scale=0.56][baseline]
\begin{axis}[
    xlabel={\bf Fraction of Retrieved Documents (Recall)},
    xmin=0, xmax=13,
    ymin=0.50, ymax=0.91,
    /pgfplots/ytick={.50,.55,...,.91},
    xtick={0,1,2,3,4,5,6,7,8,9,10, 11,12,13},
    xticklabels={0.0001, 0.0005, 0.001, 0.002, 0.005, 0.01, 0.02, 0.05, 0.1, 0.2, 0.3, 0.5, 0.8, 1.0},
    x tick label style={rotate=48,anchor=east},
    legend pos=south west,
    ymajorgrids=true, xmajorgrids=true,
    grid style=dashed,
]

\addplot[
	color=blue,
	mark=square,
	]
	plot coordinates {
    (0, 0.896)
    (1, 0.893)
    (2, 0.889)
    (3, 0.885)
    (4, 0.882)
    (5, 0.878)
    (6, 0.874)
    (7, 0.865)
    (8, 0.854)
    (9, 0.833)
    (10, 0.810)
    (11, 0.739)
    (12, 0.583)
    (13, 0.499)
	};
\addlegendentry{ctx-DocNADEe}

\addplot[
	color=red,
	mark=square,
	]
	plot coordinates {
    (0, 0.888)
    (1, 0.880)
    (2, 0.880)
    (3, 0.882)
    (4, 0.879)
    (5, 0.877)
    (6, 0.874)
    (7, 0.869)
    (8, 0.861)
    (9, 0.848)
    (10, 0.829)
    (11, 0.763)
    (12, 0.586)
    (13, 0.499)
	};
\addlegendentry{ctx-DocNADE}

\addplot[
	color=cyan,
	mark=triangle,
	]
	plot coordinates {
    (0, 0.889)
    (1, 0.872)
    (2, 0.868)
    (3, 0.864)
    (4, 0.858)
    (5, 0.854)
    (6, 0.848)
    (7, 0.838)
    (8, 0.826)
    (9, 0.804)
    (10, 0.780)
    (11, 0.713)
    (12, 0.578)
    (13, 0.499)
    
	};
\addlegendentry{DocNADE(FV)}

\addplot[
	color=black,
	mark=*,
	]
	plot coordinates {
    (0, 0.837)
    (1, 0.840)
    (2, 0.837)
    (3, 0.834)
    (4, 0.828)
    (5, 0.824)
    (6, 0.820)
    (7, 0.811)
    (8, 0.801)
    (9, 0.782)
    (10, 0.760)
    (11, 0.701)
    (12, 0.576)
     (13, 0.499)
	};
\addlegendentry{DocNADE(RV)}

\addplot[
	color=green,
	mark=*,
	]
	plot coordinates {
    (0, 0.846)
    (1, 0.831)
    (2, 0.826)
    (3, 0.820)
    (4, 0.805)
    (5, 0.791)
    (6, 0.775)
    (7, 0.744)
    (8, 0.713)
    (9, 0.668)
    (10, 0.634)
    (11, 0.581)
    (12, 0.525)
    (13, 0.499)
	};
\addlegendentry{glove(FV)}

\addplot[
	color=orange,
	mark=*,
	]
	plot coordinates {
    (0, 0.813)
    (1, 0.807)
    (2, 0.800)
    (3, 0.794)
    (4, 0.782)
    (5, 0.770)
    (6, 0.753)
    (7, 0.725)
    (8, 0.695)
    (9, 0.653)
    (10, 0.620)
    (11, 0.571)
    (12, 0.521)
    (13, 0.499)
	};
\addlegendentry{glove(RV)}

\addplot[
	color=violet,
	mark=triangle,
	]
	plot coordinates {
    (0, 0.847)
    (1, 0.844)
    (2, 0.843)
    (3, 0.843)
    (4, 0.842)
    (5, 0.841)
    (6, 0.839)
    (7, 0.836)
    (8, 0.831)
    (9, 0.822)
    (10, 0.809)
    (11, 0.750)
    (12, 0.583)
    (13, 0.499)
	};
\addlegendentry{TDLM}

\end{axis}
\end{tikzpicture}%
\caption{{\bf IR:} Subjectivity} \label{IRsubjectivity}
\end{subfigure}\hspace*{\fill}
\begin{subfigure}{0.33\textwidth}
\centering
\begin{tikzpicture}[scale=0.56][baseline]
\begin{axis}[
    xlabel={\bf Fraction of Retrieved Documents (Recall)},
    xmin=0, xmax=13,
    ymin=0.50, ymax=0.67,
   /pgfplots/ytick={.50,.52,...,.67},
    xtick={0,1,2,3,4,5,6,7,8,9,10,11,12,13},
    xticklabels={0.0001, 0.0005, 0.001, 0.002, 0.005, 0.01, 0.02, 0.05, 0.1, 0.2, 0.3, 0.5, 0.8, 1.0},
    x tick label style={rotate=45,anchor=east},
    legend pos=north east,
    ymajorgrids=true, xmajorgrids=true,
    grid style=dashed,
]

\addplot[
	color=blue,
	mark=square,
	]
	plot coordinates {
    (0, 0.660)
    (1, 0.650)
    (2, 0.642)
    (3, 0.634)
    (4, 0.624)
    (5, 0.614)
    (6, 0.603)
    (7, 0.587)
    (8, 0.575)
    (9, 0.559)
    (10, 0.549)
    (11, 0.533)
    (12, 0.513)
   (13, 0.499)
	};
\addlegendentry{ctx-DocNADEe}

\addplot[
	color=red,
	mark=square,
	]
	plot coordinates {
    (0, 0.650)
    (1, 0.636)
    (2, 0.625)
    (3, 0.619)
    (4, 0.607)
    (5, 0.599)
    (6, 0.591)
    (7, 0.578)
    (8, 0.567)
    (9, 0.553)
    (10, 0.544)
    (11, 0.530)
    (12, 0.512)
   (13, 0.499)
	};
\addlegendentry{ctx-DocNADE}

\addplot[
	color=cyan,
	mark=triangle,
	]
	plot coordinates {
    (0, 0.647)
    (1, 0.632)
    (2, 0.617)
    (3, 0.605)
    (4, 0.595)
    (5, 0.585)
    (6, 0.576)
    (7, 0.562)
    (8, 0.552)
    (9, 0.542)
    (10, 0.534)
    (11, 0.523)
    (12, 0.509)
    (13, 0.499)
	};
\addlegendentry{DocNADE(FV)}

\addplot[
	color=black,
	mark=*,
	]
	plot coordinates {
    (0, 0.643)
    (1, 0.623)
    (2, 0.610)
    (3, 0.596)
    (4, 0.581)
    (5, 0.570)
    (6, 0.560)
    (7, 0.546)
    (8, 0.537)
    (9, 0.528)
    (10, 0.523)
    (11, 0.515)
    (12, 0.506)
    (13, 0.499)
    
	};
\addlegendentry{DocNADE(RV)}

\addplot[
	color=green,
	mark=*,
	]
	plot coordinates {
    (0, 0.606)
    (1, 0.591)
    (2, 0.588)
    (3, 0.577)
    (4, 0.566)
    (5, 0.559)
    (6, 0.553)
    (7, 0.541)
    (8, 0.532)
    (9, 0.523)
    (10, 0.518)
    (11, 0.511)
    (12, 0.503)
  (13, 0.499)
	};
\addlegendentry{glove(FV)}

\addplot[
	color=orange,
	mark=*,
	]
	plot coordinates {
    (0, 0.591)
    (1, 0.585)
    (2, 0.579)
    (3, 0.568)
    (4, 0.557)
    (5, 0.550)
    (6, 0.542)
    (7, 0.534)
    (8, 0.526)
    (9, 0.519)
    (10, 0.515)
    (11, 0.509)
    (12, 0.503)
  (13, 0.499)
	};
\addlegendentry{glove(RV)}

\addplot[
	color=violet,
	mark=triangle,
	]
	plot coordinates {
    (0, 0.546)
    (1, 0.536)
    (2, 0.530)
    (3, 0.525)
    (4, 0.522)
    (5, 0.520)
    (6, 0.518)
    (7, 0.516)
    (8, 0.514)
    (9, 0.511)
    (10, 0.509)
    (11, 0.507)
    (12, 0.502)
    (13, 0.499)
	};
\addlegendentry{TDLM}

\end{axis}
\end{tikzpicture}%
\caption{{\bf IR:} Polarity} \label{IRPolarity}
\end{subfigure}\hspace*{\fill}
~%

\begin{subfigure}{0.33\textwidth}
\centering
\begin{tikzpicture}[scale=0.56][baseline]
\begin{axis}[
    xlabel={\bf Fraction of Retrieved Documents  (Recall)},
    ylabel={\bf Precision (\%)},
    xmin=0, xmax=11,
    ymin=0.22, ymax=0.76,
    /pgfplots/ytick={.25,.30,...,.76},
    xtick={0,1,2,3,4,5,6,7,8,9, 10,11,12,13},
    xticklabels={0.0001, 0.0005, 0.001, 0.002, 0.005, 0.01, 0.02, 0.05, 0.1, 0.2, 0.3, 0.5},
    x tick label style={rotate=48,anchor=east},
    legend pos=south west,
    ymajorgrids=true, xmajorgrids=true,
    grid style=dashed,
]

\addplot[
	color=blue,
	mark=square,
	]
	plot coordinates {
    (0, 0.739)
    (1, 0.698)
    (2, 0.681)
    (3, 0.665)
    (4, 0.641)
    (5, 0.620)
    (6, 0.595)
    (7, 0.549)
    (8, 0.496)
    (9, 0.417)
    (10, 0.357)
    (11, 0.273)
    (12, 0.200)
    (13, 0.169)
	};
\addlegendentry{ctx-DocNADEe}

\addplot[
	color=red,
	mark=square,
	]
	plot coordinates {
    (0, 0.719)
    (1, 0.679)
    (2, 0.660)
    (3, 0.640)
    (4, 0.612)
    (5, 0.588)
    (6, 0.560)
    (7, 0.512)
    (8, 0.463)
    (9, 0.394)
    (10, 0.341)
    (11, 0.266)
    (12, 0.199)
    (13, 0.169)
	};
\addlegendentry{ctx-DocNADE}

\addplot[
	color=cyan,
	mark=triangle,
	]
	plot coordinates {
    (0, 0.712)
    (1, 0.676)
    (2, 0.656)
    (3, 0.632)
    (4, 0.600)
    (5, 0.574)
    (6, 0.576)
    (7, 0.501)
    (8, 0.455)
    (9, 0.387)
    (10, 0.335)
    (11, 0.263)
    (12, 0.198)
    (13, 0.169)
	};
\addlegendentry{DocNADE(FV)}

\addplot[
	color=black,
	mark=*,
	]
	plot coordinates {
    (0, 0.665)
    (1, 0.634)
    (2, 0.620)
    (3, 0.600)
    (4, 0.573)
    (5, 0.549)
    (6, 0.524)
    (7, 0.480)
    (8, 0.435)
    (9, 0.372)
    (10, 0.324)
    (11, 0.259)
    (12, 0.198)
    (13, 0.169)
	};
\addlegendentry{DocNADE(RV)}

\addplot[
	color=green,
	mark=*,
	]
	plot coordinates {
    (0, 0.746)
    (1, 0.700)
    (2, 0.675)
    (3, 0.649)
    (4, 0.612)
    (5, 0.581)
    (6, 0.525)
    (7, 0.489)
    (8, 0.434)
    (9, 0.359)
    (10, 0.309)
    (11, 0.246)
    (12, 0.192)
    (13, 0.169)
	};
\addlegendentry{glove(FV)}

\addplot[
	color=orange,
	mark=*,
	]
	plot coordinates {
    (0, 0.694)
    (1, 0.653)
    (2, 0.633)
    (3, 0.610)
    (4, 0.574)
    (5, 0.544)
    (6, 0.512)
    (7, 0.461)
    (8, 0.411)
    (9, 0.343)
    (10, 0.296)
    (11, 0.239)
    (12, 0.190)
    (13, 0.169)
	};
\addlegendentry{glove(RV)}

\addplot[
	color=violet,
	mark=triangle,
	]
	plot coordinates {
    (0, 0.603)
    (1, 0.593)
    (2, 0.586)
    (3, 0.577)
    (4, 0.564)
    (5, 0.552)
    (6, 0.535)
    (7, 0.500)
    (8, 0.458)
    (9, 0.389)
    (10, 0.335)
    (11, 0.261)
    (12, 0.197)
    (13, 0.169)
	};
\addlegendentry{TDLM}

\end{axis}
\end{tikzpicture}%
\caption{{\bf IR:} TMNtitle} \label{IRTMNtitle}
\end{subfigure}\hspace*{\fill}%
~~~%
\begin{subfigure}{0.33\textwidth}
\centering
\begin{tikzpicture}[scale=0.56][baseline]
\begin{axis}[
    xlabel={\bf Fraction of Retrieved Documents (Recall)},
    xmin=0, xmax=11,
    ymin=0.30, ymax=0.81,
    /pgfplots/ytick={.30,.35,...,.81},
    xtick={0,1,2,3,4,5,6,7,8,9, 10,11,12,13},
    xticklabels={0.0001, 0.0005, 0.001, 0.002, 0.005, 0.01, 0.02, 0.05, 0.1, 0.2, 0.3, 0.5}, 
    x tick label style={rotate=48,anchor=east},
    legend pos=south west,
    legend style={font=\fontsize{7}{8}\selectfont},
    ymajorgrids=true, xmajorgrids=true,
    grid style=dashed,
]

\addplot[
	color=blue,
	mark=square,
	]
	plot coordinates {
    (0, 0.792)
    (1, 0.768)
    (2, 0.757)
    (3, 0.747)
    (4, 0.732)
    (5, 0.718)
    (6, 0.701)
    (7, 0.667)
    (8, 0.623)
    (9, 0.544)
    (10, 0.477)
    (11, 0.380)
    (12, 0.291)
    (13, 0.250)
	};
\addlegendentry{ctx-DocNADEe}

\addplot[
	color=red,
	mark=square,
	]
	plot coordinates {
    (0, 0.785)
    (1, 0.760)
    (2, 0.749)
    (3, 0.738)
    (4, 0.722)
    (5, 0.708)
    (6, 0.691)
    (7, 0.657)
    (8, 0.611)
    (9, 0.532)
    (10, 0.467)
    (11, 0.375)
    (12, 0.290)
    (13, 0.250)
	};
\addlegendentry{ctx-DocNADE}

\addplot[
	color=cyan,
	mark=triangle,
	]
	plot coordinates {
    (0, 0.789)
    (1, 0.759)
    (2, 0.746)
    (3, 0.732)
    (4, 0.713)
    (5, 0.697)
    (6, 0.678)
    (7, 0.643)
    (8, 0.598)
    (9, 0.519)
    (10, 0.457)
    (11, 0.369)
    (12, 0.288)
    (13, 0.250)
	};
\addlegendentry{DocNADE(FV)}

\addplot[
	color=black,
	mark=*,
	]
	plot coordinates {
    (0, 0.758)
    (1, 0.734)
    (2, 0.722)
    (3, 0.708)
    (4, 0.690)
    (5, 0.674)
    (6, 0.656)
    (7, 0.621)
    (8, 0.577)
    (9, 0.503)
    (10, 0.445)
    (11, 0.363)
    (12, 0.287)
    (13, 0.250)
	};
\addlegendentry{DocNADE(RV)}

\addplot[
	color=green,
	mark=*,
	]
	plot coordinates {
    (0, 0.777)
    (1, 0.738)
    (2, 0.718)
    (3, 0.698)
    (4, 0.669)
    (5, 0.643)
    (6, 0.612)
    (7, 0.558)
    (8, 0.502)
    (9, 0.429)
    (10, 0.383)
    (11, 0.324)
    (12, 0.273)
    (13, 0.250)
	};
\addlegendentry{glove(FV)}

\addplot[
	color=orange,
	mark=*,
	]
	plot coordinates {
    (0, 0.750)
    (1, 0.710)
    (2, 0.691)
    (3, 0.671)
    (4, 0.642)
    (5, 0.617)
    (6, 0.587)
    (7, 0.536)
    (8, 0.483)
    (9, 0.416)
    (10, 0.374)
    (11, 0.320)
    (12, 0.272)
    (13, 0.250)
	};
\addlegendentry{glove(RV)}

\addplot[
	color=violet,
	mark=triangle,
	]
	plot coordinates {
    (0, 0.576)
    (1, 0.569)
    (2, 0.565)
    (3, 0.560)
    (4, 0.553)
    (5, 0.545)
    (6, 0.534)
    (7, 0.511)
    (8, 0.480)
    (9, 0.428)
    (10, 0.388)
    (11, 0.332)
    (12, 0.277)
    (13, 0.250)
	};
\addlegendentry{TDLM}

\end{axis}
\end{tikzpicture}%
\caption{{\bf IR:} AGnewstitle} \label{IRAGnewstitle}
\end{subfigure}\hspace*{\fill}%
\begin{subfigure}{0.33\textwidth}
\centering
\begin{tikzpicture}[scale=0.56][baseline]
\begin{axis}[
    xlabel={\bf Fraction of Retrieved Documents (Recall)},
    xmin=0, xmax=11,
    ymin=0.10, ymax=0.71,
   /pgfplots/ytick={.10,.20,...,.71},
    xtick={0,1,2,3,4,5,6,7,8,9, 10,11,12,13},
    xticklabels={0.0001, 0.0005, 0.001, 0.002, 0.005, 0.01, 0.02, 0.05, 0.1, 0.2, 0.3, 0.5},
    x tick label style={rotate=45,anchor=east},
    legend pos=north east,
    legend style={font=\fontsize{8}{9}\selectfont},
    ymajorgrids=true, xmajorgrids=true,
    grid style=dashed,
]

\addplot[
	color=blue,
	mark=square,
	]
	plot coordinates {
    (0, 0.687)
    (1, 0.657)
    (2, 0.659)
    (3, 0.602)
    (4, 0.565)
    (5, 0.532)
    (6, 0.484)
    (7, 0.382)
    (8, 0.282)
    (9, 0.189)
    (10, 0.144)
    (11, 0.097)
    (12, 0.065)
    (13, 0.052)
	};
\addlegendentry{ctx-DocNADEe}

\addplot[
	color=red,
	mark=square,
	]
	plot coordinates {
    (0, 0.669)
    (1, 0.642)
    (2, 0.615)
    (3, 0.587)
    (4, 0.552)
    (5, 0.518)
    (6, 0.472)
    (7, 0.372)
    (8, 0.276)
    (9, 0.186)
    (10, 0.142)
    (11, 0.097)
    (12, 0.064)
    (13, 0.052)
	};
\addlegendentry{ctx-DocNADE}

\addplot[
	color=cyan,
	mark=triangle,
	]
	plot coordinates {
    (0, 0.657)
    (1, 0.618)
    (2, 0.588)
    (3, 0.557)
    (4, 0.517)
    (5, 0.478)
    (6, 0.428)
    (7, 0.336)
    (8, 0.251)
    (9, 0.174)
    (10, 0.137)
    (11, 0.096)
    (12, 0.064)
    (13, 0.052)
	};
\addlegendentry{DocNADE(FV)}

\addplot[
	color=black,
	mark=*,
	]
	plot coordinates {
    (0, 0.582)
    (1, 0.538)
    (2, 0.509)
    (3, 0.477)
    (4, 0.437)
    (5, 0.406)
    (6, 0.366)
    (7, 0.294)
    (8, 0.225)
    (9, 0.159)
    (10, 0.126)
    (11, 0.090)
    (12, 0.063)
    (13, 0.052)
	};
\addlegendentry{DocNADE(RV)}

\addplot[
	color=green,
	mark=*,
	]
	plot coordinates {
    (0, 0.524)
    (1, 0.466)
    (2, 0.428)
    (3, 0.387)
    (4, 0.335)
    (5, 0.295)
    (6, 0.253)
    (7, 0.196)
    (8, 0.155)
    (9, 0.117)
    (10, 0.098)
    (11, 0.076)
    (12, 0.059)
     (13, 0.052)
	};
\addlegendentry{glove(FV)}

\addplot[
	color=orange,
	mark=*,
	]
	plot coordinates {
    (0, 0.434)
    (1, 0.381)
    (2, 0.347)
    (3, 0.308)
    (4, 0.263)
    (5, 0.231)
    (6, 0.199)
    (7, 0.158)
    (8, 0.128)
    (9, 0.101)
    (10, 0.087)
    (11, 0.070)
    (12, 0.057)
     (13, 0.052)
	};
\addlegendentry{glove(RV)}

\addplot[
	color=violet,
	mark=triangle,
	]
	plot coordinates {
    (0, 0.532)
    (1, 0.512)
    (2, 0.501)
    (3, 0.483)
    (4, 0.455)
    (5, 0.428)
    (6, 0.387)
    (7, 0.308)
    (8, 0.232)
    (9, 0.160)
    (10, 0.125)
    (11, 0.089)
    (12, 0.063)
    (13, 0.052)
	};
\addlegendentry{ProdLDA (FV)}

\end{axis}
\end{tikzpicture}%
\caption{{\bf IR:} 20NS} \label{IR20NS}
\end{subfigure}\hspace*{\fill}
\caption{
Retrieval performance (IR-precision) on 
6 datasets at different fractions
}
\label{fig:docretrieval}
\end{figure*}
\makeatletter

\subsection{Applicability: Text Retrieval and Categorization}\label{sec:applicability}
{\it Text Retrieval}: 
We perform a document retrieval task using the short-text and long-text documents with label information.
We follow the experimental setup similar to  \citet{HugoJMLR:82}, 
where all test documents are treated as queries to retrieve a fraction of the closest documents in
the original training set using cosine similarity measure between their \texttt{textTOvec} representations (section \ref{sec:ctx-docnade}).  
To compute retrieval precision for each fraction (e.g., $0.0001$, $0.005$, $0.01$, $0.02$, $0.05$, etc.), 
we average the number of retrieved training documents with the same label as the query. 
For multi-label datasets, we average the precision scores over multiple labels for each query.  
Since,  \citet{Salakhutdinov:82}  and  \citet{HugoJMLR:82}  have 
shown that RSM and DocNADE strictly outperform LDA on this task, 
we solely compare DocNADE  with our proposed extensions.

Table \ref{PPLIRF1scoresshorttext} and \ref{PPLIRF1scoreslongtext} show the retrieval precision scores for the short-text and long-text datasets, respectively  at retrieval fraction $0.02$. 
Observe that the introduction of both pre-trained embeddings and language/contextual information leads to improved performance on the IR task noticeably for short texts.    
We also investigate topic modeling without pre-processing and filtering certain words, i.e. the FV setting and find that the DocNADE(FV) or glove(FV) improves IR precision over the baseline RV setting. 
Therefore, we opt for the FV in the proposed extensions. 
On an average over the 8 short-text and 6 long-text datasets, {\it ctx-DocNADEe} reports a gain of $7.1$\% ($.630$ vs $.588$) (Table \ref{PPLIRF1scoresshorttext}) 
 $6.0$\% ($.601$ vs $.567$) (Table \ref{PPLIRF1scoreslongtext}), respectively in precision compared to DocNADE(RV). 
To further compare with TDLM, our proposed models (ctx-DocNADE and ctx-DocNADEe) outperform it by a notable margin for all the short-text datasets, i.e., a gain of 14.5\% (.630 vs .550: ctx-DocNADEe vs TDLM) in IR-precision. 
In addition, the deep variant ($d$=3) with embeddings, i.e., ctx-DeepDNEe shows competitive performance on TREC6 and Subjectivity datasets. 

Figures (\ref{IR20NSshort},  \ref{IRsubjectivity},  \ref{IRPolarity}, \ref{IRTMNtitle}, \ref{IRAGnewstitle} and \ref{IR20NS}) illustrate   
the average precision for the retrieval task on 6 datasets. Observe that the ctx-DocNADEe outperforms DocNADE(RV) at all the fractions and demonstrates 
a gain of $6.5$\%  ($.615$ vs $.577$) in precision at fraction 0.02, averaged over 14 datasets.   
Additionally, our proposed models outperform TDLM and ProdLDA\footnote{IR-precision scores at 0.02 retrieval fraction on the short-text datasets by ProdLDA: {\it 20NSshort} (.08),  {\it TREC6} (.24), {\it R21578title} (.31), {\it Subjectivity} (.63) and {\it Polarity} (.51).  Therefore, the {\it DocNADE, ctx-DocNADE and ctx-DocNADEe outperform ProdLDA in both the settings}: data sparsity and sufficient co-occurrences.} (for 20NS) by noticeable margins.  

{\it Text Categorization}:  We perform text categorization to  measure the quality of our \texttt{textTovec} representations.  
We consider the same experimental setup as in the retrieval task and extract \texttt{textTOvec} of 200 dimension for each document,  learned during the training of ctx-DocNADE variants.  
To perform text categorization, we employ 
a logistic regression classifier 
with $L2$ regularization.  
While, {\it ctx-DocNADEe} and {\it ctx-DeepDNEe} make use of glove embeddings, they are evaluated against the topic model baselines with embeddings. 
For the short texts (Table \ref{PPLIRF1scoresshorttext}), the {\it glove} leads DocNADE in classification performance, suggesting a need for distributional priors in the topic model. 
Therefore, the ctx-DocNADEe reports a gain of  $4.8$\% ($.705$ vs $.673$)  and $3.6$\%($.618$ vs $.596$) in $F1$, compared to DocNADE(RV)
 on an average over the short (Table \ref{PPLIRF1scoresshorttext}) and long (Table \ref{PPLIRF1scoreslongtext}) texts, respectively.  
In result, a gain of  $4.4$\%  ($.662$ vs $.634$) overall. 

In terms of classification accuracy on 20NS dataset, the scores are: 
DocNADE (0.734), 
ctx-DocNADE (0.744), 
ctx-DocNADEe (0.751), 
NTM (0.72) and 
SCHOLAR (0.71). 
While, our proposed models, i.e., ctx-DocNADE and ctx-DocNADEe outperform both NTM 
(results taken from \cite{cao2015novel}, Figure 2) and SCHOLAR (results taken from \cite{card2017neural}, Table 2), 
the DocNADE establishes itself as a strong neural topic model baseline.  

\subsection{Inspection of Learned Representations}
To further interpret the topic models, we analyze the meaningful semantics captured via topic extraction.  
Table \ref{topiccoherenceexamples} shows a topic extracted using 20NS dataset that could be interpreted as {\it computers}, 
which are (sub)categories in the data, confirming that meaningful topics are captured. 
Observe that the ctx-DocNADEe extracts a more coherent topic due to embedding priors. 
To {\it qualitatively} inspect the contribution of word embeddings and \texttt{textTOvec} representations in topic models, 
we analyse the text retrieved for each query using the representations learned from DocNADE and ctx-DoocNADEe models. 
Table \ref{exampleIR} illustrates the retrieval of the top 3 texts for an input query, selected from {\it TMNtitle} dataset, 
where \#match is YES if the query and retrievals have the same class label.   
Observe that ctx-DocNADEe retrieves the top 3 texts, each with no unigram overlap with the query.

\begin{minipage}{0.44\linewidth}
\begin{table}[H]
\centering
\renewcommand*{\arraystretch}{1.25}
\resizebox{1.0\textwidth}{!}{
\setlength\tabcolsep{2.5pt}
\begin{tabular}{c|c|c}
{\it DocNADE}   &     {\it ctx-DocNADE}        & {\it ctx-DocNADEe}           \\ \hline
    vga, screen,                    &                 computer, color,                  &                         svga, graphics                       \\
   computer, sell,                      &         screen,  offer,                         &         bar, macintosh,                                     \\
    color,    powerbook,                 &       vga, card,                            &           san, windows,                                 \\
    sold, cars,                    &                        terminal, forsale,         &            utility, monitor,                                      \\
    svga, offer                    &                   gov, vesa               &                computer,  processor              \\ \hline 
         .554     &    .624     &    \underline{.667}                           
\end{tabular}}
\caption{A topic of 20NS dataset with coherence}
\label{topiccoherenceexamples}
\end{table}
\end{minipage}\hfill
\begin{minipage}{0.55\linewidth}
\begin{table}[H]
\centering
\small
\renewcommand*{\arraystretch}{1.3}
\resizebox{0.99\textwidth}{!}{
\setlength\tabcolsep{2.5pt}
\begin{tabular}{rrrrl|c}
 &   \multirow{3}{*}{\rotatebox{90}{-DocNADEe}} & {\it Query}  & :: & ``emerging economies move ahead nuclear plans"    &  \#match \\ \cline{3-6}\cline{3-6}
\multirow{3}{*}{\rotatebox{90}{ctx-}}   &     &  \#IR1      &::  &  imf sign lifting japan yen                                          &    YES\\
&   &    \#IR2        &::  &  japan recovery takes hold debt downgrade looms     &  YES\\
&   &   \#IR3        &::  &  japan ministers confident treasuries move              &   YES \\ \cline{3-6}
   & \multirow{3}{*}{\rotatebox{90}{DocNADE}}  &    \#IR1          &::  & nuclear regulator back power plans                         &      NO   \\ 
&    &    \#IR2         &::  &  defiant iran plans big rise nuclear                          &     NO \\
&    &   \#IR3        &::  &  japan banks billion nuclear operator sources          &    YES  
\end{tabular}}
\caption{Illustration of the top-3 retrievals for an input query}
\label{exampleIR}
\end{table}
\end{minipage}

\begin{figure*}[t]
\centering
\begin{subfigure}{0.49\textwidth}
\centering
\begin{tikzpicture}[scale=0.75][baseline]
\begin{axis}[
    xlabel={\bf Fraction of training set},
    ylabel={\bf IR-precision at fraction 0.02},
     xmin=-0.3, xmax=4.3,
    ymin=0.42, ymax=0.61,
   /pgfplots/ytick={.42,.44,...,.61},
    xtick={0, 1, 2, 3, 4},
    xticklabels={20\%, 40\%, 60\%, 80\%, 100\%},
    legend pos=south east,
    ymajorgrids=true, xmajorgrids=true,
    grid style=dashed,
]

\addplot[
	color=blue,
	mark=square,
	]
	plot coordinates {
    (0, 0.580)
    (1, 0.585)
    (2, 0.588)
    (3, 0.590) 
   (4, 0.595)
	};
\addlegendentry{ctx-DocNADEe}

\addplot[
	color=red,
	mark=square,
	]
	plot coordinates {
    (0, 0.539)
    (1, 0.544)
    (2, 0.549)
    (3, 0.551) 
	(4, 0.560)
	};
\addlegendentry{ctx-DocNADE}

\addplot[
	color=cyan,
	mark=triangle,
	]
	plot coordinates {
    (0, 0.444)
    (1, 0.498)
    (2, 0.513)
    (3, 0.522) 
	(4, 0.525)
    
	};
\addlegendentry{DocNADE(FV)}

\end{axis}
\end{tikzpicture}%
\caption{Text Retrieval} \label{IRprecision}
\end{subfigure}\hspace*{\fill}%
~%
\begin{subfigure}{0.49\textwidth}
\centering
\begin{tikzpicture}[scale=0.75][baseline]
\begin{axis}[
    xlabel={\bf Fraction of training set},
    ylabel={\bf macro-F1 score},
     xmin=-0.3, xmax=4.3,
    ymin=0.605, ymax=0.74,
   /pgfplots/ytick={.61,.62,...,.73},
    xtick={0, 1, 2, 3, 4},
    xticklabels={20\%, 40\%, 60\%, 80\%, 100\%},
    legend pos=south east,
    ymajorgrids=true, xmajorgrids=true,
    grid style=dashed,
]

\addplot[
	color=blue,
	mark=square,
	]
	plot coordinates {
    (0, 0.711)
    (1, 0.719)
    (2, 0.724)
    (3, 0.721) 
	(4, 0.726)
	};
\addlegendentry{ctx-DocNADEe}

\addplot[
	color=red,
	mark=square,
	]
	plot coordinates {
    (0, 0.678)
    (1, 0.682)
    (2, 0.687)
    (3, 0.685) 
	(4, 0.687)
	};
\addlegendentry{ctx-DocNADE}

\addplot[
	color=cyan,
	mark=triangle,
	]
	plot coordinates {
    (0, 0.615)
    (1, 0.655)
    (2, 0.666)
    (3, 0.681) 
	(4, 0.688)
    
	};
\addlegendentry{DocNADE(FV)}

\end{axis}
\end{tikzpicture}%
\caption{Text classification} \label{textretrieval}
\end{subfigure}\hspace*{\fill}%
\caption{Evaluations at different fractions (20\%, 40\%, 60\%, 80\%, 100\%) of the training set of TMNtitle}
\label{fig:differenttrainingfractions}
\end{figure*}
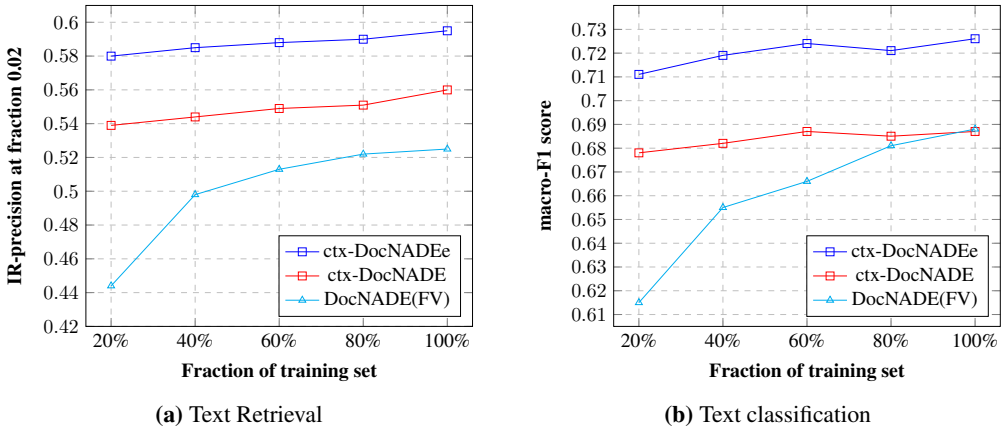

Additionally, we show the quality of representations learned at different fractions (20\%, 40\%, 60\%, 80\%, 100\%) of training set from TMNtitle data and use the same experimental setup for the IR and classification tasks, as in section \ref{sec:applicability}.  
In Figure \ref{fig:differenttrainingfractions}, we quantify the quality of representations learned and demonstrate improvements due to the proposed models, i.e., ctx-DocNADE and ctx-DocNADEe over DocNADE at different fractions of the training data. 
Observe that the gains in both the tasks are large for smaller fractions of the datasets. For instance, one of the proposed models, i.e., ctx-DocNADEe (vs DocNADE) reports: 
(1) a precision (at 0.02 fraction) of 0.580 vs 0.444 at 20\% and 0.595 vs 0.525 at 100\% of the training set, and 
(2) an F1 of 0.711 vs 0.615 at 20\% and 0.726 vs 0.688 at 100\% of the training set. Therefore, the findings conform to our second contribution of improving topic models with word embeddings, especially in the sparse data setting.

\subsection{Conclusion}
In this work, we have shown that accounting for language concepts such as  word ordering, syntactic and semantic information in neural autoregressive  topic models 
helps to better estimate the probability of a word in a given context. To this end, we have combined a neural autoregressive topic- (i.e., DocNADE) and a neural language (e.g., LSTM-LM) model in a single probabilistic framework with an aim to introduce language concepts in each of the autoregressive steps of the topic model. 
This facilitates learning a latent representation from the entire document whilst accounting for the local dynamics of the collocation patterns, encoded in the internal states of LSTM-LM. We further augment this complementary learning with external knowledge by introducing word embeddings. 
Our experimental results show that our proposed modeling approaches consistently outperform state-of-the-art generative topic models, quantified by generalization (perplexity), topic interpretability (coherence), and applicability (text retrieval and categorization) on 15 datasets.

\makeatletter
\def\labelonly{BDF}
\def\labelcheck#1{
    \edef\pgfmarshal{\noexpand\pgfutil@in@{#1}{\labelonly}}
    \pgfmarshal
    \ifpgfutil@in@[#1]\fi
}
\makeatother

\makeatletter

\bibliography{iclr2019_conference}
\bibliographystyle{iclr2019_conference}

\appendix

\section{Data Description}

We use 14 different datasets: 
(1) \texttt{20NSshort}: We take documents from 20NewsGroups data, with document size less (in terms of number of words) than 20. 
(2) \texttt{TREC6}: a set of questions 
(3) \texttt{Reuters21578title}:  a collection of new stories from \url{nltk.corpus}. We take titles of the documents.    
(4) \texttt{Subjectivity}: sentiment analysis data. 
(5) \texttt{Polarity}: a collection of positive and negative snippets acquired from Rotten Tomatoes 
(6)  \texttt{TMNtitle}: Titles of the Tag My News (TMN) news dataset. 
(7) \texttt{AGnewstitle}: Titles of the AGnews dataset. 
(8) \texttt{Reuters8}: a collection of news stories, processed and released by 
(9) \texttt{Reuters21578}:  a collection of new stories from \url{nltk.corpus}.   
(10) \texttt{20NewsGroups}: a collection of  news stories from \url{nltk.corpus}.   
(11) RCV1V2 (Reuters): {\url{www.ai.mit.edu/projects/jmlr/papers/volume5/lewis04a/lyrl2004_rcv1v2_README.htm}}  
(12) \texttt{20NSsmall}: We sample 20 document for training from each class of the 20NS dataset. For validation and test, 10 document for each class. 
(13) \texttt{TMN}: The Tag My News (TMN) news dataset.
(14) \texttt{Sixxx Requirement OBjects} (\texttt{SiROBs}): a collection of paragraphs extracted from industrial tender documents (our industrial corpus). 

The SiROBs is our industrial corpus, extracted from industrial tender documents. 
The documents contain requirement specifications for an industrial project for example, {\it railway metro construction}. 
There are 22 types of requirements i.e. class labels (multi-class), where a requirement is a paragraph or collection of paragraphs within a document. 
We name the requirement as Requirement Objects (ROBs). Some of the requirement types are {\it project management}, {\it testing}, {\it legal}, {\it risk analysis}, {\it financial cost}, {\it technical requirement}, etc. 
We need to classify the requirements in the tender documents and assign each ROB to a relevant department(s).  Therefore,  we analyze such documents to automate decision making, 
tender comparison, similar tender as well as ROB retrieval and assigning ROBs to a relevant 
department(s) to optimize/expedite tender analysis. See some examples of ROBs from SiROBs corpus in Table \ref{SiROBsexamples}.

\begin{table*}[t]
\centering
\renewcommand*{\arraystretch}{1.2}
\resizebox{0.99\textwidth}{!}{
\begin{tabular}{|c|}
\hline 
{\bf Label:} {\it training}              \\ \hline
Instructors shall have tertiary education and experience in the 
operation and maintenance \\
of the equipment or sub-system of Plant. They shall be proficient 
in the use of the English language both written \\
and oral. They shall be able to deliver instructions clearly and
 systematically. The curriculum vitae \\
of the instructors shall be submitted for acceptance by the 
Engineer at least 8 weeks before \\
the commencement of any training.  \\ \hline \hline

{\bf Label:}   {\it maintenance}    \\ \hline
 The Contractor shall provide experienced staff for 24 hours per Day, 
7 Days per week, throughout the Year, \\
for call out to carry out On-call  Maintenance for the Signalling System.  \\ \hline \hline

{\bf Label:}   {\it cables} \\ \hline
Unless otherwise specified, this standard is applicable to all cables 
which include single and multi-core cables \\ 
and wires, Local Area Network  (LAN) cables and Fibre Optic (FO) cables.  \\ \hline \hline

{\bf Label:}   {\it installation} \\ \hline
The Contractor shall provide and permanently install the asset labels onto 
all equipment supplied \\
under this Contract. The Contractor shall liaise and  co-ordinate with 
the Engineer for the format \\
and the content of the labels. The Contractor shall submit the final format
 and size of the labels as well \\ 
as the installation layout of the labels on the respective equipment, to 
the Engineer for acceptance.\\ \hline \hline 

{\bf Label:}   {\it operations, interlocking} \\ \hline
It shall be possible to switch any station Interlocking capable of reversing 
the service into  \\
``Auto-Turnaround Operation". This facility once selected shall automatically 
route Trains into and out  of \\
these stations, independently of the ATS system. At stations where multiple
 platforms can be used  to reverse\\
 the service it shall be possible to select one or both platforms for the service reversal. \\ \hline \hline
\end{tabular}}
\caption{SiROBs data: Example Documents (Requirement Objects) with their types (label).}
\label{SiROBsexamples}
\end{table*}

\section{Experimental Setup}

\subsection{Experimental Setup and Hyperparameters for Generalization task}
See Table \ref{HyperparametersinGeneralization}  for hyperparameters used in generalization.

\begin{table}[t]
      \centering
        \begin{tabular}{c|c}
         \hline 
         {\bf Hyperparameter}               & {\bf Search Space} \\ \hline
           learning rate        &    [0.001]                   \\
           hidden units        &       [200]                      \\
           iterations        &      [2000]      \\
           activation function      &    sigmoid            \\ 
          $\lambda$        &      [1.0, 0.8, 0.5, 0.3, 0.1, 0.01, 0.001]  \\ \hline      
       \end{tabular}
         \caption{Hyperparameters in Generalization in the DocNADE and ctx-DocNADE variants for 200 topics}\label{HyperparametersinGeneralization}
 \end{table}%

\subsection{Experimental Setup and Hyperparameters for IR task}

We set the maximum number of training passes to 1000, topics to 200 and the learning rate to 0.001 with $tanh$ hidden activation.  
For model selection, we used the validation set as the query set and used the average precision at 0.02 retrieved documents as the
performance measure.  
Note that the labels are not used during training. The class labels are only used to check if the retrieved documents have the same class label as the query document. 
To perform document retrieval, we use the same train/development/test split of documents discussed in data statistics (experimental section) for 
all the datasets during learning. 
 
See Table \ref{appendixHyperparametersinIR} for the hyperparameters in the document retrieval task.

\begin{table}[t]
      \centering
       \begin{tabular}{c|c}
        \hline 
        {\bf Hyperparameter}               & {\bf Search Space} \\ \hline
          retrieval fraction        &    [0.02]                        \\
          learning rate        &    [{0.001}]                       \\
          hidden units         &      [200]               \\ 
          activation function        &      {tanh}     \\
          iterations        &      [2000]      \\
          $\lambda$        &      [1.0, 0.8, 0.5, 0.3, 0.1, 0.01, 0.001]  \\ \hline 
         \end{tabular}
 \caption{Hyperparameters in the Document Retrieval task.}
\label{appendixHyperparametersinIR}
\end{table}

\subsection{Experimental Setup for doc2vec model}

We used gensim (\url{https://github.com/RaRe-Technologies/gensim}) to train Doc2Vec models for 12 datasets. Models were trained with distributed bag of words, for 1000 iterations using a window size of 5 and a vector size of 500.
\subsection{Classification task}
We used the same split in training/development/test as for training the Doc2Vec models (also same split as in IR task) and trained a regularized logistic regression classifier on the inferred document vectors to predict class labels. In the case of multilabel datasets (\texttt{R21578},\texttt{R21578title}, \texttt{RCV1V2}), we used a one-vs-all approach. Models were trained with a liblinear solver using L2 regularization and accuracy and macro-averaged F1 score were computed on the test set to quantify predictive power. 

\begin{table*}[t]
\center
\renewcommand*{\arraystretch}{1.2}
\resizebox{.6\textwidth}{!}{
\begin{tabular}{c|c||ccc}
\hline
\multicolumn{1}{c}{Dataset}   & \multicolumn{1}{c}{Model} & \multicolumn{3}{c}{\bf $\lambda$}        \\
                     &    & 1.0    & 0.1    & 0.01      \\ \hline

\multirow{2}{*}{20NSshort}    & ctx-DocNADE               & 899.04 & 829.5  &  842.1            \\
                              & ctx-DocNADEe              & 890.3  & 828.8  & 832.4          \\  \hline

\multirow{2}{*}{Subjectivity} & ctx-DocNADE               & 982.8      & 977.8  & 966.5     \\
                              & ctx-DocNADEe              & 977.1      & 975.0 & 964.2        \\  \hline
\multirow{2}{*}{TMNtitle}     & ctx-DocNADE               & 1898.1 & 1482.7 & 1487.1        \\
                              & ctx-DocNADEe              & 1877.7 & 1480.2 & 1484.7        \\  \hline
\multirow{2}{*}{AGnewstitle}  & ctx-DocNADE               & 1296.1 &  861.1     & 865           \\
                              & ctx-DocNADEe              & 1279.2 & 853.3      & 862.9       \\ \hline
\multirow{2}{*}{Reuters-8}    & ctx-DocNADE               & 336.1  & 313.2    & 311.9        \\
                              & ctx-DocNADEe              & 323.3  & 312.0     & 310.2        \\ \hline
\multirow{2}{*}{20NS}         & ctx-DocNADE               & 1282.1 & 1209.3 & 1207.2      \\
                              & ctx-DocNADEe              & 1247.1 & 1211.6      & 1206.1       

\end{tabular}}
\caption{Perplexity scores for different $\lambda$ in Generalization task: Ablation over validation set}
\label{lambdappl}
\end{table*}

\begin{table}[t]
\center
\renewcommand*{\arraystretch}{1.2}
\resizebox{.60\textwidth}{!}{
\begin{tabular}{l|l|llll}
\hline
\multicolumn{1}{c}{Dataset}   & \multicolumn{1}{c}{Model} & \multicolumn{4}{c}{\bf $\lambda$}     \\ \hline
                              &                           & 1.0   & 0.8   & 0.5   & 0.3   \\
\multirow{2}{*}{20NSshort}    & ctx-DocNADE               & 0.264 & 0.265 & 0.265 & 0.265 \\
                              & ctx-DocNADEe              & 0.277 & 0.277 & 0.278 & 0.276 \\ \hline
\multirow{2}{*}{Subjectivity} & ctx-DocNADE               & 0.874 & 0.874 & 0.873 & 0.874 \\
                              & ctx-DocNADEe              & 0.868 & 0.868 & 0.874 & 0.87  \\ \hline
\multirow{2}{*}{Polarity}     & ctx-DocNADE               & 0.587 & 0.588 & 0.591 & 0.587 \\
                              & ctx-DocNADEe              & 0.602 & 0.603 & 0.601 & 0.599 \\ \hline
\multirow{2}{*}{TMNtitle}     & ctx-DocNADE               & 0.556 & 0.557 & 0.559 & 0.568 \\
                              & ctx-DocNADEe              & 0.604 & 0.604 & 0.6   & 0.6   \\ \hline
\multirow{2}{*}{TMN}          & ctx-DocNADE               & 0.683 & 0.689 & 0.692 & 0.694 \\  
                              & ctx-DocNADEe              & 0.696 & 0.698 & 0.698 & 0.7   \\ \hline
\multirow{2}{*}{AGnewstitle}  & ctx-DocNADE               & 0.665 & 0.668 & 0.678 & 0.689 \\
                              & ctx-DocNADEe              & 0.686 & 0.688 & 0.695 & 0.696 \\  \hline
\multirow{2}{*}{20NSsmall}    & ctx-DocNADE               & 0.352 & 0.356 & 0.366 & 0.37  \\
                              & ctx-DocNADEe              & 0.381 & 0.381 & 0.375 & 0.353 \\  \hline
\multirow{2}{*}{Reuters-8}    & ctx-DocNADE               &    0.863   &  0.866     & 0.87  & 0.87  \\
                              & ctx-DocNADEe              & 0.875      & 0.872 & 0.873 &   0.872    \\  \hline
\multirow{2}{*}{20NS}         & ctx-DocNADE               & 0.503 & 0.506 & 0.513 & 0.512 \\
                              & ctx-DocNADEe              & 0.524 & 0.521 & 0.518 & 0.511 \\  \hline
\multirow{2}{*}{R21578}       & ctx-DocNADE               & 0.714 & 0.714 & 0.714 & 0.714 \\
                              & ctx-DocNADEe              & 0.715 & 0.715 & 0.715 & 0.714 \\   \hline
\multirow{2}{*}{SiROBs}       & ctx-DocNADE               & 0.409 & 0.409 & 0.408 & 0.408 \\
                              & ctx-DocNADEe              & 0.41  & 0.411 & 0.411 & 0.409 \\  \hline
\multirow{2}{*}{AGnews}       & ctx-DocNADE               & 0.786 & 0.789 & 0.792 & 0.797 \\
                              & ctx-DocNADEe              & 0.795 & 0.796 & 0.8   & 0.799
\end{tabular}}
\caption{$\lambda$ for IR task: Ablation over validation set at retrieval fraction 0.02}
\label{lambdIR}
\end{table}

\subsection{Experimental Setup for glove-DMM and glove-LDA  models}
We used LFTM (\url{https://github.com/datquocnguyen/LFTM}) to train glove-DMM and glove-LDA models. Models were trained for 200 iterations with 2000 initial iterations using 200 topics. For short texts we set the hyperparameter beta to 0.1, for long texts to 0.01; the mixture parameter lambda was set to 0.6 for all datasets.
The setup for the classification task was the same as for doc2vec; classification was performed using relative topic proportions as input (i.e. we inferred the topic distribution of the training and test documents and used the relative distribution as input for the logistic regression classifier). Similarly, for the IR task, similarities were computed based on the inferred relative topic distribution.

\subsection{Experimental Setup for ProdLDA}
We run ProdLDA (\url{https://github.com/akashgit/autoencoding_vi_for_topic_models}) on the short-text datasets in the FV setting to generate document vectors for IR-task. We use 200 topics for a fair comparison with other baselines used for the IR tasks.  We infer topic distribution of the training and test documents and used the relative distribution as input for the IR task, similar to section 3.3.

To fairly compare PPL scores of ProdLDA and DocNADE in the RV setting, we take the preprocessed 20NS dataset released by ProdLDA and run DocNADE for 200 topics. To further compare them in the FV setting, we run ProdLDA (\url{https://github.com/akashgit/autoencoding_vi_for_topic_models}) on the processed 20NS dataset for 200 topics used in this paper.

\section{Ablation over the mixture weight $\lambda$}

\subsection{$\lambda$ for Generalization task}
See Table \ref{lambdappl}. 

\subsection{$\lambda$ for IR task}
See Table \ref{lambdIR}.  

\section{Additional Baselines}

\subsection{DocNADE vs SCHOLAR}
{\it PPL scores over 20 topics}: DocNADE (752) and SCHOLAR (921), i.e., DocNADE outperforms SCHOLAR in terms of generalization. 

{\it Topic coherence (NPMI) using 20 topics}: DocNADE (.18) and SCHOLAR (.35), i.e., SCHOLAR  \citep{card2017neural} generates more coherence topics than DocNADE, though worse in PPL and text classification (see section 3.3) than DocNADE, ctx-DocNADE and ctx-DocNADEe. 

{\it IR tasks}: Since, SCHOLAR  \citep{card2017neural} without meta-data equates to ProdLDA and we have shown in section 3.3 that ProdLDA is worse on IR tasks than our proposed models,  therefore one can infer the performance of SCHOLAR on IR task.  

The experimental results above suggest that the DocNADE is better than SCHOLAR in generating good representations for downstream tasks such as information retrieval or classification, however falls behind SCHOLAR in interpretability. The investigation opens up an interesting direction for future research.  

\end{document}